\documentclass[a4paper,11pt]{article}
\advance\hoffset by -0.5truein
\usepackage[utf8]{inputenc}
\usepackage{graphicx}
\usepackage{epstopdf}
\usepackage{diagbox}
\usepackage{tikz}
\newtheorem{tw}{Theorem}
\newtheorem{ex}{Example}

\usepackage{amsmath}
\usepackage{amssymb}

\usepackage{inputenc}
\usepackage{xcolor}
\usepackage{url}
\usepackage{pdflscape}

\usepackage{amssymb,latexsym}

\usepackage{enumerate}

 \usepackage[T1]{fontenc}

\newfont{\mm}{eufm10 scaled 1200}
\newtheorem{obs}{Observation}

\newtheorem{corol}{Corollary}

\newtheorem{define}{Definition}
\newtheorem{implication}{Implication}
\begin{document}

%\newpage
\centerline{\Large\bf Empirical Evaluation of No Free Lunch Violations}
\vspace{0.2cm}
\centerline{\Large\bf  in Permutation-Based Optimization}
\vspace{0.3cm}

\centerline{\large Grzegorz Sroka } \vspace{4mm}
\centerline{\footnotesize\it Department of Nonlinear Analysis, 
Rzesz\'ow University of Technology, Powsta\'nc\'ow Warszawy 12,}
\centerline{\footnotesize\it  35-959 Rzesz\'ow, Poland}
\centerline{\footnotesize\it  e-mail: gsroka@prz.edu.pl}
\vspace{0,25cm}

{\footnotesize  {\bf Abstract.}  \it  \  The No Free Lunch (NFL) theorem guarantees equal average performance only under uniform sampling of a function space closed under permutation (c.u.p.).  We ask when this averaging ceases to reflect what benchmarking actually reports. We study an iterative-search setting with sampling without replacement, where algorithms differ only in evaluation order. Binary objectives allow exhaustive evaluation in the fully enumerable case, and efficiency is defined by the first time the global minimum is reached. We then construct two additional benchmarks by algebraically recombining the same baseline functions through sums and differences. Function-algorithm relations are examined via correlation structure, hierarchical clustering, delta heatmaps, and PCA. A one-way ANOVA with Tukey contrasts confirms that algebraic reformulations induce statistically meaningful shifts in performance patterns. The uniformly sampled baseline remains consistent with the global NFL symmetry. In contrast, the algebraically modified benchmarks yield stable re-rankings and coherent clusters of functions and sampling policies. Composite objectives can also exhibit non-additive search effort despite being built from simpler components. Monte Carlo experiments indicate that order effects persist in larger spaces and depend on function class. Taken together, the results show how objective reformulation and benchmark design can generate structured local departures from NFL intuition. They motivate algorithm choice that is aware of both the problem class and the objective representation. This message applies to evolutionary computation as well as to statistical procedures based on relabeling, resampling, and permutation tests.}

\vskip0.15cm

\section{Introduction}

Optimization in practice is performed on finite, discrete machines.
Under these conditions, the NFL theorem can be a poor guide.
Performance may depend on structure in a restricted problem class, not on averages over the full space of all functions.

Is there a universal optimization algorithm that performs equally well across all possible problems? 
This question lies at the heart of one of the most intriguing results in optimization  -  the NFL theorem. 
Introduced in 1995, NFL suggests that, on average, all optimization algorithms perform identically when evaluated over the entire space of possible problems. 
This provocative claim has sparked significant debate and criticism.  

In his book \cite{5}, Goldberg argued that traditional optimization methods are effective only for a narrow class of problems, while robust techniques like evolutionary algorithms achieve good results across a wider range. 
However, the authors of a seminal 1995 report \cite{12ass} countered this, stating that no algorithm outperforms another when averaged over all possible problems.  
The NFL theorem has also been used in broader disputes. 
For instance, Dembski \cite{4} invoked the NFL framework in arguments related to intelligent design. 

As Clerc observes \cite{3}, computation is finite and discrete. 
This remark is the direct inspiration for our study. 
This makes it natural to study restricted classes of functions, including those that are not closed under permutation (not-c.u.p.):
\begin{quote}
On a computer everything is discrete finite so we consider more carefully
not-c.u.p. sets of functions. It appears that except for very particular cases,
not only the NFL does not hold, not only there is a best algorithm in
average, but that an ''ultimate'' algorithm does exist.
\end{quote}
This perspective motivates our focus on structured subsets where NFL-style averaging may fail and where performance differences among algorithms can persist.

This paper aims to revisit the NFL theorem, focusing on specific conditions under which it does not hold. 
By analyzing subsets of binary functions closed under permutation (c.u.p.) and arithmetic operations, we demonstrate that certain algorithmic groups exhibit differing levels of efficiency, challenging the universality of NFL.

In this study, we focus on permutation-based optimization algorithms due to their broad applicability in combinatorial optimization problems such as the traveling salesman problem, scheduling, assignment problems, and routing. 
These algorithms exploit symmetries in the solution space and are widely used in evolutionary computation and heuristic optimization. 
Because permutation spaces are finite and discrete, they provide a well-defined framework for examining theoretical implications of NFL in settings close to practice.

To conduct a controlled and exhaustive study of algorithm performance across the full space of binary objective functions, we restrict our analysis to the case $n=4$, where all $4!=24$ permutations and the corresponding $2^{24}$ binary functions can be fully enumerated. 
This dimensionality allows exact measurement of algorithm efficiency without sampling error and enables precise verification of structural properties such as c.u.p. 
For larger dimensions, exhaustive analysis becomes infeasible due to the rapid growth of the function space. 
For $n=5$, the space grows from $2^{24}$ to $2^{120}$, i.e., by a factor of $2^{96} \approx 10^{28}$. 
For $n=10$, the number of permutations is $10!=3628800$, and the full function space contains $2^{10!}=2^{3628800}$ binary functions, which is out of reach even for HPC systems.

It should be noted that even for $n=4$ the computations can be tedious. 
For much larger domains, such as $2^{30}$, $2^{50}$, or $2^{100}$, manual investigation is practically impossible and would require substantial computational resources, both in processing power and memory. 
In such cases, a more effective approach is to apply statistical methods and analyze selected binary samples with specific characteristics. 
For example, one could select functions whose output vectors are highly imbalanced, e.g., where at least $91\%$ of all bits are equal to either `'1'' or ``0''. 
One could also select sequences in which $91\%$ of the initial positions in the binary string are ``1''. 
Such sampling rules could facilitate the study of higher dimensions and improve the efficiency of the analysis.

Hence, $n=4$ offers a balance between tractability and analytical rigor, making it suitable for uncovering conditions under which the assumptions of the NFL theorem may no longer hold. 
Extension to higher dimensions using statistical sampling is planned for future research and lies outside the scope of this work.

The methodological choices in our study were guided by the aim of ensuring clarity, rigor, and direct interpretability of the results. 
Restricting the codomain to $\{0,1\}$ was an intentional analytical simplification, enabling full enumeration of the function space (for small problem sizes) and isolation of structural properties relevant to the NFL theorem. 
This follows a well-established practice in the NFL literature (e.g., Ho and Pepyne \cite{7}), where binary functions serve as minimal yet representative models. 
Importantly, the conclusions extend to broader codomains, as the observed effects depend on problem structure rather than on value range. 
Likewise, the assumption of sampling without replacement reflects the original NFL framework (Wolpert and Macready \cite{12ass}). 
It ensures that each evaluation step provides new information and allows fair comparison of information-efficient strategies. 
Although idealized, this assumption is consistent with mechanisms present in many practical metaheuristics (e.g., tabu search, duplicate elimination in evolutionary algorithms), where revisiting solutions is avoided or minimized. 
By adopting these design decisions, we eliminate artifacts unrelated to problem structure and focus on essential factors influencing algorithm performance. 
Our findings confirm that when the problem class departs from the fully random case, structural regularities can systematically favor certain algorithms, even under the most information-efficient conditions.

\paragraph{Methodology.}
This study analyzed the efficiency of 24 permutation-based algorithms on a set of 16 binary functions closed under permutation (c.u.p.). 
Statistical analyses, including correlation matrices, hierarchical clustering, and PCA, were applied to assess the impact of algebraic transformations of functions on optimization. 
The study identified conditions under which the NFL does not hold, highlighting the role of function structure in the varying effectiveness of optimization algorithms.

\section{Related work}
We briefly review recent advances in permutation optimization that exploit the geometry, algebra, and probabilistic structure of permutation spaces.

Recent research on permutation problems emphasizes the necessity of tailoring operators and models to the specific structure of this search space. Moraglio and Poli \cite{100bb} introduced the concept of geometric crossover for permutation representation, based on a defined metric of the solution space and unifying existing recombination operators within a common geometric framework. Santucci et al. \cite{10ccg},\cite{10cch} proposed an algebraic approach, treating the set of permutations as a group structure, which enabled the transfer of continuous algorithms (e.g., differential evolution and particle swarm optimization) into the discrete domain by defining corresponding operations on permutations. From a probabilistic perspective, Ceberio et al. \cite{2cvb} analyzed the impact of the choice of permutation distance metric (e.g., Kendall, Cayley, Ulam) in Mallows and Generalized Mallows models within EDAs, showing that an appropriate distance measure significantly affects algorithmic performance and yields smoother fitness landscapes. Santucci and Ceberio \cite{10sfj} introduced a doubly stochastic matrix (DSM) model for EDAs tackling the Quadratic Assignment Problem (QAP), which more effectively captures solution structure and outperforms earlier probabilistic models, while Ceberio and Santucci \cite{2abu} proposed a model-based gradient search method that reduces computational overhead and improves scalability for permutation problems. Doerr et al. \cite{4ytg} conducted a formal runtime analysis of evolutionary algorithms in permutation spaces, adapting classical benchmarks (such as \textit{Leading Ones} and \textit{Jump}) to this domain, and demonstrated that mutation difficulty depends not only on Hamming distance but also on the cycle structure of a given permutation. These works - encompassing generalized metric spaces, probabilistic models, geometric operators, and algebraic formalisms - are consistent with the approach adopted in this study or represent its natural extension, confirming that the proposed method aligns with current research directions in permutation optimization algorithms.

Collectively, these studies frame our controlled evaluation of permutation-based algorithms on structured benchmark classes.

\section{Notation}

Let $\mathcal{X}$ be a discrete set of states and $\mathcal{Y}$ be a discrete set of values that can be assigned
these states by mapping~$f\colon \mathcal{X} \to \mathcal{Y}$. We assume that both sets are finite, and - in the context of optimization - $f$ is the cost function, also called objective function, energy function, etc. A specific function $f$ corresponds to a certain optimization problem, while the set~$\mathcal{F} = \mathcal{Y}^\mathcal{X}$ is the space of all possible optimization problems. In particular, when $\mathcal{Y}=\{0, 1\}$, the set $\mathcal{F}$ consists of $2^{|\mathcal{X}|}$ binary functions.

\vspace{0.1cm}
\begin{ex}
Let the set $\mathcal{X} = \{1, 2, 3, 4\}$ and $\mathcal{Y} = \{0, 1\}$ denote the set of values. Then $|\mathcal{X}|=4.$ $|\mathcal{Y}|=2$ and we have $|\mathcal{F}|=m^{n}=2^4=16$ possible functions: $\mathcal{X}\to\mathcal{Y}$ defined by their values on the set $\mathcal{X}$ (see Table\ref{EVO:tab-3}).
\end{ex}
In this example, we fix $n = 4$ to denote the input dimension, i.e., the number of elements in the domain $\mathcal{X}$, and $m = 2$ to denote the cardinality of the codomain $\mathcal{Y}$.
The functions are fully specified by their output values on each element of $\mathcal{X}$, and the input strings in Table~\ref{EVO:tab-3} represent all binary combinations of length $n = 4$.

\begin{center}
\begin{table}[h]
%\tiny{
\centering
\scalebox{0.9}{
\begin{tabular}{|c||c|c|c|c|c|c|c|c|c|c|c|c|c|c|c|c|}
 \hline
\diagbox[width=3.5cm]{$f\in\mathcal{F}$}
{Values on  $\mathcal{X}$}
& $f1$ & $f2$ & $f3$ & $f4$ & $f5$ & $f6$ & $f7$ & $f8$ &$f9$& $f10$ & $f11$ & $f12$ & $f13$ & $f14$& $f15$ & $f16$\\
 \hline $x_1$ & 0 & 1 & 0 & 1 & 0 & 1 & 0 & 1 & 0 & 1 & 0 & 1 & 0 & 1 & 0 & 1\\
 \hline $x_2$ & 0 & 0 & 1 & 1 & 0 & 0 & 1 & 1 & 0 & 0 & 1 & 1 & 0 & 0 & 1 & 1\\
 \hline $x_3$ & 0 & 0 & 0 & 0 & 1 & 1 & 1 & 1 & 0 & 0 & 0 & 0 & 1 & 1 & 1 & 1\\
 \hline $x_4$ & 0 & 0 & 0 & 0 & 0 & 0 & 0 & 0 & 1 & 1 & 1 & 1 & 1 & 1 & 1 & 1\\
 %\hline Średnia & 0.67 & 1.0 & 0.5 & 1.0 & 0.67 & 0.5\\
 \hline
 \end{tabular}
 }
 \caption{16 preparation functions.}
\label{EVO:tab-3}
\end{table}
\end{center}
The strings used in the above table should be understood as follows: for each function $f \in \mathcal{F}$ we have: the values of the function $fi$ are binary expansions of the number $i - 1$, written in reverse bit order (i.e., least significant bit first).
For example, $f4(1)=1$, $f4(2)=1$, $f4(3)=0$, $f4(4)=0$. A wide range of applications of binary functions has been discussed, among others, in papers: \cite{7}, \cite{10b}, \cite{12a}, \cite{12aa}.

%Evaluating the efficiency of the algorithm ~$f$ to solve the problem means evaluating number of functions necessary to find the given problem. 

We measure the efficiency of an algorithm designed to solve a problem $f$ by the number of objective-function evaluations required to reach a prescribed solution quality. We assume that \textit{different} points of space ~$\mathcal{X}$ are evaluated. The result of the search is the sequence

\[T_m=\{<x_1,f(x_1)>,\ <x_2,f(x_2)>,\dots,<x_m,f(x_m)>\}\]
where $1\leq x_{i}\leq m$, for $m\in\{1,...,|\mathcal{X}|\}$ ($|\mathcal{X}|$ here means the size of the set $\mathcal{X}$),  $x_{i}\in\mathcal{X}$ and $f(x_i)=y_{i}\in\mathcal{Y}$
\cite{5}, \cite{11}.

In order to define the concept of an optimization algorithm, we will introduce definitions regarding the sampling principle.

When optimizing a function $f:\mathcal{X}\to\mathcal{Y}$ one can assume that the sets $\mathcal{X}$ and $\mathcal{Y}$ are known, but only a few specific values $y=f(x)$ are known. Then that partial information about $f$ can be described by a function $d : \mathcal{X}\to\mathcal{Y}\cup\{?\}$ and is called a \textit{data}. Symbol "$?$" means an unknown  value of  $f(x).$ For all $x$,
for which we know the values $y=f(x)$ we get $d(x)=f(x),$ for the remaining $x$ we get $d(x)= \ ?$.

Let $\mathcal{D}$ be the set of all possible ''data'' for the function in $\mathcal{F}$. Hence $|\mathcal{D}|=(m+1)^n$ and for each $x\in\mathcal{X}$, $d(x)= \ ?$ this means "no data", i.e. situation, in which we know nothing about the cost function.

\begin{define}
\textbf{(Sampling Policy)} \cite{10} A sampling policy $s$ is a function $s: \mathcal{D}\to\mathcal{X}$, where $\mathcal{D}$ is the set of all possible data (defined above). 
\end{define}
\begin{define}
\textbf{(Non-Repeating Policy)} \cite{10} A non-repeating sampling policy $s$ is a function  $s: \mathcal{D}\to\mathcal{X}$ s.t. if $s(d)=x$ then either $d(x)= \ ?$ or $|d|=m$.
\end{define}
\begin{define}
\textbf{(Optimization Algorithm)} \cite{10}
An optimization algorithm $A$ based on sampling policy $s$ is iterated use of that sampling policy and data updating:
\vspace{-0.15cm}
\begin{itemize}
\item [(a)]
Set  $d$ to be initial data,
\vspace{-0.25cm}
\item [(b)] 
If there are no unsampled $x\in\mathcal{X}$, then terminate,
\vspace{-0.25cm}
\item [(c)] 
Sample at $x=s(d)$ and add the result to the data $d,$ i.e. set $d(x)=f(x)$,
\vspace{-0.25cm}
\item [(d)] 
Go to step (b).
\end{itemize}
\end{define}
\vspace{0.1cm}

We present an exemplary optimization algorithm using the directed graph shown in Figure \ref{EVO:graph-27A}. In this graph, nodes represent data. Edges show all potential data transitions resulting from sampling the problem function according to the optimizer’s sampling principle. Figure \ref{EVO:graph-27A} is inspired by \cite{10}.

The paper is organized as follows. Section~4 introduces the permutational iterative-search framework and the benchmarking protocol, including the efficiency measure. Sections~5-8 analyze algorithm-function dependencies and the effects of algebraic structure and transformations. Section~9 discusses implications for statistics and evolutionary computing and concludes the paper.

 \usetikzlibrary{arrows,trees,positioning}

\tikzset{
  treenode/.style = {align=center, inner sep=0pt, text centered,
    font=\sffamily},
  arn_n/.style = {treenode, circle, white, font=\sffamily\bfseries, draw=black,
    fill=black, text width=0.8em},% arbre rouge noir, noeud noir
  arn_r/.style = {treenode, circle, red, draw=red, 
    text width=1.5em, very thick},% arbre rouge noir, noeud rouge
  arn_x/.style = {treenode, rectangle, draw=black,
    minimum width=2.5em, minimum height=1em}% arbre rouge noir, nil
}

\begin{figure}
\scalebox{0.47}{
\begin{tikzpicture}[->,>=stealth',level/.style={sibling distance = 25cm/#1,
  level distance = 1.5cm}]
%\scalebox{0.6}{
\node [arn_r] {\tiny{????}}
    child{node [arn_x] {\tiny{s(????)=1}}
    child{node [arn_r, xshift=-2cm, yshift=0.5cm] {\tiny{??0?}} child{node [arn_x] {\tiny{s(??0?)=2}} child{node [arn_r, xshift=-1cm, yshift=1cm] {\tiny{?00?}}child{node [arn_x] {\tiny{s(?00?)=3}}child{node [arn_r] {\tiny{000?}}
    child{node [arn_x] {\tiny{s(000?)=4}}child{node [arn_r] {\tiny{0000}} edge from parent node[above left]
  {\tiny{$f(4)=0$}}}
   child{node [arn_r] {\tiny{0001}} edge from parent node[above right, yshift=-0.3cm]
 {\tiny{$f(4)=1$}}}}
     edge from parent node[above left]
  {\tiny{$f(3)=0$}}}
   child{node [arn_r] {\tiny{100?}}
   child{node [arn_x] {\tiny{s(100?)=4}}child{node [arn_r] {\tiny{1000}} edge from parent node[above left]
  {\tiny{$f(4)=0$}}}
   child{node [arn_r] {\tiny{1001}} edge from parent node[above right, yshift=-0.3cm]
 {\tiny{$f(4)=1$}}}} edge from parent node[above right]
 {\tiny{$f(3)=1$}}}} edge from parent node[above left]
 {\tiny{$f(2)=0$}}  }  child{node [arn_r, xshift=1cm, yshift=1cm] {\tiny{?10?}} child{node [arn_x] {\tiny{s(?10?)=3}} child{node [arn_r] {\tiny{010?}}
 child{node [arn_x] {\tiny{s(010?)=4}}child{node [arn_r] {\tiny{0100}} edge from parent node[above left]
  {\tiny{$f(4)=0$}}}
   child{node [arn_r] {\tiny{0101}} edge from parent node[above right, yshift=-0.3cm]
 {\tiny{$f(4)=1$}}}}
 edge from parent node[above left]
 {\tiny{$f(3)=0$}} } child{node [arn_r] {\tiny{011?}}
 child{node [arn_x] {\tiny{s(011?)=4}}child{node [arn_r] {\tiny{0110}} edge from parent node[above left]
  {\tiny{$f(4)=0$}}}
   child{node [arn_r] {\tiny{0111}} edge from parent node[above right, yshift=-0.3cm]
 {\tiny{$f(4)=1$}}}}
  edge from parent node[above right]
 {\tiny{$f(3)=1$}} } } edge from parent node[above right]
 {\tiny{$f(2)=1$}} }} edge from parent node[above left]
 {\tiny{$f(1)=0$}}}
child{node [arn_r, xshift=2cm, yshift=0.5cm] {\tiny{??1?}}child{node [arn_x]{\tiny{s(??1?)=2}} child{node [arn_r, xshift=-1cm, yshift=1cm] {\tiny{??10}}child{node [arn_x] {\tiny{s(??10)=3}} child{node [arn_r] {\tiny{0?10}}
child{node [arn_x] {\tiny{s(0?10)=4}}child{node [arn_r] {\tiny{0010}} edge from parent node[above left]
  {\tiny{$f(4)=0$}}}
   child{node [arn_r] {\tiny{0110}} 
   edge from parent node[above right, yshift=-0.3cm]
 {\tiny{$f(4)=1$}}}}
 edge from parent node[above left]
 {\tiny{$f(3)=0$}}} child{node [arn_r] {\tiny{1?10}}
 child{node [arn_x] {\tiny{s(1?10)=4}}child{node [arn_r] {\tiny{1010}} edge from parent node[above left]
  {\tiny{$f(4)=0$}}}
   child{node [arn_r] {\tiny{1110}} edge from parent node[above right, yshift=-0.3cm]
 {\tiny{$f(4)=1$}}}}
 edge from parent node[above right]
 {\tiny{$f(3)=1$}}}} edge from parent node[above left]
 {\tiny{$f(2)=0$}} } child{node [arn_r, xshift=1cm, yshift=1cm] {\tiny{??11}} child{node [arn_x] {\tiny{s(??11)=3}} child{node [arn_r] {\tiny{0?11}} 
 child{node [arn_x] {\tiny{s(0?11)=4}}child{node [arn_r] {\tiny{0011}} edge from parent node[above left]
  {\tiny{$f(4)=0$}}}
   child{node [arn_r] {\tiny{0111}} edge from parent node[above right, yshift=-0.3cm]
 {\tiny{$f(4)=1$}}}}
 edge from parent node[above left]
 {\tiny{$f(3)=0$}}} child{node [arn_r] {\tiny{1?11}} 
 child{node [arn_x] {\tiny{s(1?11)=4}}child{node [arn_r] {\tiny{1011}} edge from parent node[above left]
  {\tiny{$f(4)=0$}}}
   child{node [arn_r] {\tiny{1111}} edge from parent node[above right, yshift=-0.3cm]
 {\tiny{$f(4)=1$}}}}
 edge from parent node[above right]
 {\tiny{$f(3)=1$}}}} edge from parent node[above right]
 {\tiny{$f(2)=1$}}  }} edge from parent node[above right]
 {\tiny{$f(1)=1$}} }
};

\end{tikzpicture}
}
\caption{Example of policies represented as a tree: a policy for functions $f:\{1, 2, 3, 4\}\to\{0, 1\}$.}
\label{EVO:graph-27A}
\end{figure}
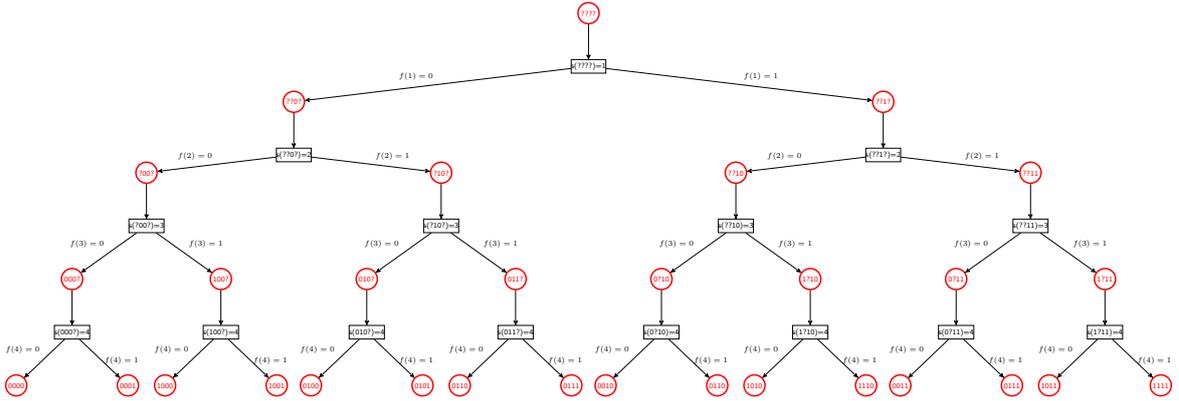

\section{Iterative search algorithm}

We define an iterative search algorithm as a sequence of sampled points with no repetitions \textit{(if we are practitioners, we probably see the algorithm as a process.
Just imagine that the definition given here is the result of such a process)}.
These are permutations (In \cite{9}:Function $\phi: \mathcal{X}\to \mathcal{X}$ we call \textit{permutation} when $\phi$ is a bijection)
positions and we have e.g. $4!=1\cdot 2\cdot 3\cdot 4=24$ possible algorithms, listed in Table \ref{EVO:tab-24-compact}.
%We denote the set of these algorithms $A$.

\vspace{-0.01cm}
Based on the above description of the point sampling problem, we can create a table of algorithms:

\begin{table}[h]
\centering
\tiny
\renewcommand{\arraystretch}{1.2}
\scalebox{0.85}{
\begin{tabular}{|c|c|c|c|c|c|c|c|c|c|c|c|}
\hline
Algorithm & $a1$ & $a2$ & $a3$ & $a4$ & $a5$ & $a6$ & $a7$ & $a8$ & $a9$ & $a10$ & $a11$  \\
\hline
Sequence & (1,2,3,4) & (1,2,4,3) & (1,3,2,4) & (1,3,4,2) & (1,4,2,3) & (1,4,3,2) & (2,1,3,4) & (2,1,4,3) & (2,3,1,4) & (2,3,4,1) & (2,4,1,3) \\
\hline
Algorithm & $a12$ & $a13$ & $a14$ & $a15$ & $a16$ & $a17$ & $a18$ & $a19$ & $a20$ & $a21$ & $a22$  \\
\hline
Sequence & (2,4,3,1) & (3,1,2,4) & (3,1,4,2) & (3,2,1,4) & (3,2,4,1) & (3,4,1,2) & (3,4,2,1) & (4,1,2,3) & (4,1,3,2) & (4,2,3,1) & (4,2,1,3) \\
\hline
Algorithm & $a23$ & $a24$ & & & & & & & & &  \\
\hline
Sequence & (4,3,1,2) & (4,3,2,1) & & & & & & & & &  \\
\hline
\end{tabular}
}
\caption{24 preparatory sequences.}
\label{EVO:tab-24-compact}
\end{table}

\vspace{0.3cm}
\subsection{Characteristics of the 24 Permutation-Based Algorithms}

All 24 algorithms presented in Table \ref{EVO:tab-24-compact}  are based on a shared strategy for exploring the solution space, differing solely in the permutation order of four predefined input points. The operational mechanism of each heuristic remains identical: the algorithms follow a deterministic, fixed sequence of four evaluations without employing any adaptive parameter tuning, randomness, or memory. In practice, this means that the only distinction between algorithms lies in the order in which they evaluate the same four points. Despite the absence of learning or memory, each algorithm encounters different results at different stages of the search process - the order of access determines which pieces of information (i.e., objective function values) are seen "earlier" and which are "delayed".
Each algorithm is conceptually defined as a deterministic procedure over the same four input configurations, without any code-based implementation, randomness, memory, or adaptation.

The sampling order directly affects the balance between exploration and exploitation. Initial evaluation points determine which regions of the input space are examined first. Algorithms that begin by sampling points close to one another (neighboring in the input space) immediately concentrate their evaluations on a limited area (favoring exploitation), whereas those starting from more distant or extreme points initially perform a broader, more distributed exploration (favoring exploration). For example, an algorithm that evaluates points in the order $(1,2,3,4)$ will quickly uncover local structures of the objective function, while a permutation such as $(1,4,2,3)$ initially surveys two boundary regions, gaining a broader view of the space at the expense of slower convergence toward promising regions.

The main consequences of these strategies can be summarized as follows:\\
\begin{itemize}
\item[•] Early region discovery: The sampling order determines which areas of the objective function are discovered first. Algorithms starting with neighboring points tend to focus on local regions (stronger exploitation), while those beginning with distant points cover a broader area (stronger exploration). As a result, different permutations yield different speeds in identifying valuable regions.\\

\item[•] No adaptation - diverse observation sequences: All algorithms in Table \ref{EVO:tab-24-compact}  are deterministic and non-adaptive: a fixed a priori evaluation order reveals regions in a different order, but never uses observed function values to choose the next point. The differing order means that each permutation generates a distinct sequence of observed function values (i.e., varying patterns of success and failure). This yields different observed-value (best-so-far) trajectories and thus different effectiveness. Under the same evaluation budget, the outcome still depends on the chosen order.

\item[•] Pattern similarity and clustering: Algorithms that share a similar initial scheme (e.g., those starting from the same point or initially evaluating neighboring elements) tend to exhibit similar behavior and performance. Closely related permutations result in similar evaluation trajectories, which manifest as clusters in correlation and hierarchical clustering analyses. These structured differences in sampling order provide a clear explanation for the observed grouping of algorithmic behaviors.\\
\end{itemize}
Although all algorithms evaluate the same four points, differences in sampling order lead to distinct informational trajectories, resulting in varying levels of effectiveness. The presented set provides a precisely controlled experimental framework for analyzing how heuristic structure influences optimization performance.

\subsection{Evaluation on Benchmark Functions}

To empirically examine how the differences among the 24 permutation-based algorithms affect their performance, we now turn to a controlled benchmark setting.
We consider the test functions $B = \{f1,...,f16\}$.
As for many cases of artificial sets of test functions, we assume that we know the minimum value of all test function problems,
here equal to zero. 
For each pair $(a_i, f_j)$, we record the smallest number of trials $s_{i,j}$ needed by algorithm $a_i$ to find the minimum of function $f_j$.
To facilitate comparison across algorithms and functions, we compute a normalized index called \textit{efficiency} \cite{3}:
\begin{equation}
E_{a_i,f_j} = \frac{|\mathcal{X}| - s_{i,j}}{|\mathcal{X}| - 1}
\end{equation}
where $|\mathcal{X}|$ denotes the number of points in the domain. For example, in our main setting with $n=4$, we have $\mathcal{X} = \{1,2,3,4\}$ and thus $|\mathcal{X}| = 4$.
Efficiency scores range from 1 (when the optimum is found in the first trial) to 0 (when it is found in the last possible trial).

\begin{define}
\textbf{(Function Permutation)}\cite{10}:
Let $\phi$ be a permutation $\phi: \mathcal{X}\to\mathcal{X},$ and let $f:\mathcal{X}\to\mathcal{Y}$ be an arbitrary  function. We call $f_{\phi}$ a function permutation where we define $f_{\phi}(x)=f(\phi(x)).$ 
\end{define}
\begin{define}
\textbf{(c.u.p)}
\cite{8}, \cite{10}, \cite{10aa}:
Let $\mathcal{F}$ be a set  function: $\mathcal{X}\to \mathcal{Y}$. We say $\mathcal{F}$ is closed under permutation, or  c.u.p., iff for any permutation $\phi: \mathcal{X}\to \mathcal{X}$, $f\in \mathcal{F} \Longrightarrow f_{\phi}\in \mathcal{F}.$
\end{define}

\begin{ex} \textbf{(not-c.u.p.)}
Let $f:=f_2$ from Table~\ref{EVO:tab-3}, i.e., $(f(1),f(2),f(3),f(4))=(1,0,0,0)$.
Let $\phi =(1\,2)$ be the transposition.
Then $(f_\phi(1),f_\phi(2),f_\phi(3),f_\phi(4))=(0,1,0,0)\neq(1,0,0,0)$,
hence $\mathcal{F}=\{f\}$ is not c.u.p.
\end{ex}

\begin{define}
\textbf{(Permutation Closure)}
\cite{9}, \cite{10}:
Let $\mathcal{F}$ be a set of function mapping: $\mathcal{X}\to \mathcal{Y}$. We define $\mathcal{F}_{c.u.p}$ as the smallest set containing $\mathcal{F}$ that is closed under permutation.
\end{define}

Not so long ago Maurice Clerc at \cite{3} proved that NFL theorem
\cite{1},
\cite{2},
\cite{4a},
\cite{8a},
\cite{7},
\cite{8},
\cite{9},
\cite{10},
\cite{10x},
\cite{10aa},
\cite{16ro},
\cite{18x},
\cite{19x},
\cite{22gs},
\cite{11},
\cite{11abc}, 
\cite{28x},
\cite{30dw}
does not occur for certain classes of continuous functions or for certain non-closed sets due to the permutation of elements (not-c.u.p.).
Some properties of functions on c.u.p. subsets can be proved. 

To visualize the performance of all 24 sampling algorithms across the test function set, we present two summary tables. 

Table~\ref{tab22} shows the number of steps $s_{i,j}$ required by algorithm $a_i$ to find the minimum of function $f_j$ (i.e., the number of evaluations before the optimum is reached for the first time). 

Table~\ref{tab23} reports the corresponding \textit{efficiency values} $E_{a_i,f_j}$ as defined by Equation~(1), normalized to the range $[0,1]$. These scores represent how early in the sampling
sequence the optimum was discovered - with 1 indicating success at the first step, and 0 indicating the last possible step.

The last row in each table reports the total number of steps or the average efficiency for each algorithm across all tested functions.

\begin{table}[htb]
\centering
\scalebox{0.61}{
 \begin{tabular}{|c||c|c|c|c|c|c|c|c|c|c|c|c|c|c|c|c|c|c|c|c|c|c|c|c|}\hline
\diagbox[width=2.4cm]{\tiny{Function}}
{\tiny{Algorithm}} &$a_1$& $a_2$ & $a_3$ & $a_4$ & $a_5$ & $a_6$ & $a_7$ & $a_8$ & $a_9$ & $a_{10}$ & $a_{11}$ & $a_{12}$ & $a_{13}$ & $a_{14}$ & $a_{15}$ & $a_{16}$ & $a_{17}$ & $a_{18}$ & $a_{19}$ & $a_{20}$ & $a_{21}$ & $a_{22}$ & $a_{23}$ & $a_{24}$\\
 \hline $f_1$ &  1 & 1 & 1 & 1 & 1 & 1 & 1 & 1 & 1 & 1 & 1 & 1 & 1 & 1 & 1 & 1 & 1 & 1 & 1 & 1 & 1 & 1 & 1 & 1\\
\hline $f_2$ &  2 & 2 & 2 & 2 & 2 & 2 & 1 & 1 & 1 & 1 & 1 & 1 & 1 & 1 & 1 & 1 & 1 & 1 & 1 & 1 & 1 & 1 & 1 & 1\\
\hline $f_3$ &  1 & 1 & 1 & 1 & 1 & 1 & 2 & 2 & 2 & 2 & 2 & 2 & 1 & 1 & 1 & 1 & 1 & 1 & 1 & 1 & 1 & 1 & 1 & 1\\
\hline $f_4$ &  3 & 3 & 2 & 2 & 2 & 2 & 3 & 3 & 2 & 2 & 2 & 2 & 1 & 1 & 1 & 1 & 1 & 1 & 1 & 1 & 1 & 1 & 1 & 1\\
\hline $f_5$ &  1 & 1 & 1 & 1 & 1 & 1 & 1 & 1 & 1 & 1 & 1 & 1 & 2 & 2 & 2 & 2 & 2 & 2 & 1 & 1 & 1 & 1 & 1 & 1\\
\hline $f_6$ &  2 & 2 & 3 & 3 & 2 & 2 & 1 & 1 & 1 & 1 & 1 & 1 & 3 & 3 & 2 & 2 & 2 & 2 & 1 & 1 & 1 & 1 & 1 & 1 \\
\hline $f_7$ &  1 & 1 & 1 & 1 & 1 & 1 & 2 & 2 & 3 & 3 & 2 & 2 & 2 & 2 & 3 & 3 & 2 & 2 & 1 & 1 & 1 & 1 & 1 & 1\\
 \hline $f_8$ &  4 & 3 & 4 & 3 & 2 & 2 & 4 & 3 & 4 & 3 & 2 & 2 & 4 & 3 & 4 & 3 & 2 & 2 & 1 & 1 & 1 & 1 & 1 & 1\\
 \hline $f_9$ &  1 & 1 & 1 & 1 & 1 & 1 & 1 & 1 & 1 & 1 & 1 & 1 & 1 & 1 & 1 & 1 & 1 & 1 & 2 & 2 & 2 & 2 & 2 & 2\\
  \hline $f_{10}$ &  2 & 2 & 2 & 2 & 3 & 3 & 1 & 1 & 1 & 1 & 1 & 1 & 1 & 1 & 1 & 1 & 1 & 1 & 3 & 3 & 2 & 2 & 2 & 2\\
   \hline $f_{11}$ &  1 & 1 & 1 & 1 & 1 & 1 & 2 & 2 & 2 & 2 & 3 & 3 & 1 & 1 & 1 & 1 & 1 & 1 & 2 & 2 & 3 & 3 & 2 & 2\\
    \hline $f_{12}$ &  3 & 4 & 2 & 2 & 4 & 3 & 3 & 4 & 2 & 2 & 4 & 3 & 1 & 1 & 1 & 1 & 1 & 1 & 4 & 3 & 3 & 4 & 2 & 2\\
  \hline $f_{13}$ &  1 & 1 & 1 & 1 & 1 & 1 & 1 & 1 & 1 & 1 & 1 & 1 & 2 & 2 & 2 & 2 & 3 & 3 & 2 & 2 & 2 & 2 & 3 & 3\\ 
   \hline $f_{14}$ &  2 & 2 & 3 & 4 & 3 & 4 & 1 & 1 & 1 & 1 & 1 & 1 & 3 & 4 & 2 & 2 & 4 & 3 & 3 & 4 & 2 & 2 & 4 & 3\\ 
    \hline $f_{15}$ &  1 & 1 & 1 & 1 & 1 & 1 & 2 & 2 & 3 & 4 & 3 & 4 & 2 & 2 & 3 & 4 & 3 & 4 & 2 & 2 & 4 & 3 & 3 & 4\\ 
     \hline $f_{16}$ &  4 & 4 & 4 & 4 & 4 & 4 & 4 & 4 & 4 & 4 & 4 & 4 & 4 & 4 & 4 & 4 & 4 & 4 & 4 & 4 & 4 & 4 & 4 & 4\\
 \hline \textbf{Total} &  \textbf{30} & \textbf{30} & \textbf{30} & \textbf{30} & \textbf{30} & \textbf{30} & \textbf{30} & \textbf{30} & \textbf{30} & \textbf{30} & \textbf{30} & \textbf{30} & \textbf{30} & \textbf{30} & \textbf{30} & \textbf{30} & \textbf{30} & \textbf{30} & \textbf{30} & \textbf{30} & \textbf{30} & \textbf{30} & \textbf{30} & \textbf{30}\\ 
%\hline \textbf{Łącznie} & 7 & 7 & 8 & 8 & 9 & 9\\
 \hline
 \end{tabular}%\qquad
 }
 \caption{Number of needed sampled points table for n=4.}
 \label{tab22}
\end{table}
%\end{center}
\begin{table}[htb]
\centering
\scalebox{0.484}{
 \begin{tabular}
 {|c||c|c|c|c|c|c|c|c|c|c|c|c|c|c|c|c|c|c|c|c|c|c|c|c|}
 \hline\diagbox[width=2.2cm]{\tiny{Funkcion}}
{\tiny{Algorithm}} &$a_1$& $a_2$ & $a_3$ & $a_4$ & $a_5$ & $a_6$ & $a_7$ & $a_8$ & $a_9$ & $a_{10}$ & $a_{11}$ & $a_{12}$ & $a_{13}$ & $a_{14}$ & $a_{15}$ & $a_{16}$ & $a_{17}$ & $a_{18}$ & $a_{19}$ & $a_{20}$ & $a_{21}$ & $a_{22}$ & $a_{23}$ & $a_{24}$\\
 \hline $f_1$ &  1 & 1 & 1 & 1 & 1 & 1 & 1 & 1 & 1 & 1 & 1 & 1 & 1 & 1 & 1 & 1 & 1 & 1 & 1 & 1 & 1 & 1 & 1 & 1\\
\hline $f_2$ &  2/3 & 2/3 & 2/3 & 2/3 & 2/3 & 2/3 & 1 & 1 & 1 & 1 & 1 & 1 & 1 & 1 & 1 & 1 & 1 & 1 & 1 & 1 & 1 & 1 & 1 & 1\\
\hline $f_3$ &  1 & 1 & 1 & 1 & 1 & 1 & 2/3 & 2/3 & 2/3 & 2/3 & 2/3 & 2/3 & 1 & 1 & 1 & 1 & 1 & 1 & 1 & 1 & 1 & 1 & 1 & 1\\
\hline $f_4$ &  1/3 & 1/3 & 2/3 & 2/3 & 2/3 & 2/3 & 1/3 & 1/3 & 2/3 & 2/3 & 2/3 & 2/3 & 1 & 1 & 1 & 1 & 1 & 1 & 1 & 1 & 1 & 1 & 1 & 1\\
\hline $f_5$ &  1 & 1 & 1 & 1 & 1 & 1 & 1 & 1 & 1 & 1 & 1 & 1 & 2/3 & 2/3 & 2/3 & 2/3 & 2/3 & 2/3 & 1 & 1 & 1 & 1 & 1 & 1\\
\hline $f_6$ &  2/3 & 2/3 & 1/3 & 1/3 & 2/3 & 2/3 & 1 & 1 & 1 & 1 & 1 & 1 & 1/3 & 1/3 & 2/3 & 2/3 & 2/3 & 2/3 & 1 & 1 & 1 & 1 & 1 & 1 \\
\hline $f_7$ &  1 & 1 & 1 & 1 & 1 & 1 & 2/3 & 2/3 & 1/3 & 1/3 & 2/3 & 2/3 & 2/3 & 2/3 & 2/3 & 2/3 & 2/3 & 2/3 & 1 & 1 & 1 & 1 & 1 & 1\\
 \hline $f_8$ &  0 & 1/3 & 0 & 1/3 & 2/3 & 2/3 & 0 & 1/3 & 0 & 1/3 & 2/3 & 2/3 & 0 & 1/3 & 0 & 1/3 & 2/3 & 2/3 & 1 & 1 & 1 & 1 & 1 & 1\\
 \hline $f_9$ &  1 & 1 & 1 & 1 & 1 & 1 & 1 & 1 & 1 & 1 & 1 & 1 & 1 & 1 & 1 & 1 & 1 & 1 & 2/3 & 2/3 & 2/3 & 2/3 & 2/3 & 2/3\\
  \hline $f_{10}$ &  2/3 & 2/3 & 2/3 & 2/3 & 1/3 & 1/3 & 1 & 1 & 1 & 1 & 1 & 1 & 1 & 1 & 1 & 1 & 1 & 1 & 1/3 & 1/3 & 2/3 & 2/3 & 2/3 & 2/3\\
   \hline $f_{11}$ &  1 & 1 & 1 & 1 & 1 & 1 & 2/3 & 2/3 & 2/3 & 2/3 & 1/3 & 1/3 & 1 & 1 & 1 & 1 & 1 & 1 & 2/3 & 2/3 & 1/3 & 1/3 & 2/3 & 2/3\\
    \hline $f_{12}$ &  1/3 & 0 & 2/3 & 2/3 & 0 & 1/3 & 1/3 & 0 & 2/3 & 2/3 & 0 & 1/3 & 1 & 1 & 1 & 1 & 1 & 1 & 0 & 1/3 & 1/3 & 0 & 2/3 & 2/3\\
  \hline $f_{13}$ &  1 & 1 & 1 & 1 & 1 & 1 & 1 & 1 & 1 & 1 & 1 & 1 & 2/3 & 2/3 & 2/3 & 2/3 & 1/3 & 1/3 & 2/3 & 2/3 & 2/3 & 2/3 & 1/3 & 1/3\\ 
   \hline $f_{14}$ &  2/3 & 2/3 & 1/3 & 0 & 1/3 & 0 & 1 & 1 & 1 & 1 & 1 & 1 & 1/3 & 0 & 2/3 & 2/3 & 0 & 1/3 & 1/3 & 0 & 2/3 & 2/3 & 0 & 1/3\\ 
    \hline $f_{15}$ &  1 & 1 & 1 & 1 & 1 & 1 & 2/3 & 2/3 & 1/3 & 0 & 1/3 & 0 & 2/3 & 2/3 & 1/3 & 0 & 1/3 & 0 & 2/3 & 2/3 & 0 & 1/3 & 1/3 & 0\\ 
     \hline $f_{16}$ &  0 & 0 & 0 & 0 & 0 & 0 & 0 & 0 & 0 & 0 & 0 & 0 & 0 & 0 & 0 & 0 & 0 & 0 & 0 & 0 & 0 & 0 & 0 & 0\\
 \hline \textbf{Mean} &  \textbf{0.71} & \textbf{0.71} & \textbf{0.71} & \textbf{0.71} & \textbf{0.71} & \textbf{0.71} & \textbf{0.71} & \textbf{0.71} & \textbf{0.71} & \textbf{0.71} & \textbf{0.71} & \textbf{0.71} & \textbf{0.71} & \textbf{0.71} & \textbf{0.71} & \textbf{0.71} & \textbf{0.71} & \textbf{0.71} & \textbf{0.71} & \textbf{0.71} & \textbf{0.71} & \textbf{0.71} & \textbf{0.71} & \textbf{0.71}\\    
%\hline \textbf{Łącznie} & 7 & 7 & 8 & 8 & 9 & 9\\
 \hline
 \end{tabular}%\qquad
 }
\caption{Efficiency algorithms table.}
\label{tab23}
\end{table}
For comparison, we also examined smaller dimensions $n=2$ and $n=3$, which require 7 and 15 trials, respectively, for full function identification.
In line with this, the mean efficiency of algorithms increases with $n$: it is $0.5$ for $n=2$, approximately $0.625$ for $n=3$, and $0.71$ for $n=4$. This trend supports the general observation that greater input dimensionality improves the discriminatory power of algorithm performance metrics.

%\end{document}

\section{Statistical and visual analysis of optimization algorithms for binary functions}

\begin{figure}[htb]
\centering
    \includegraphics[scale=0.39]{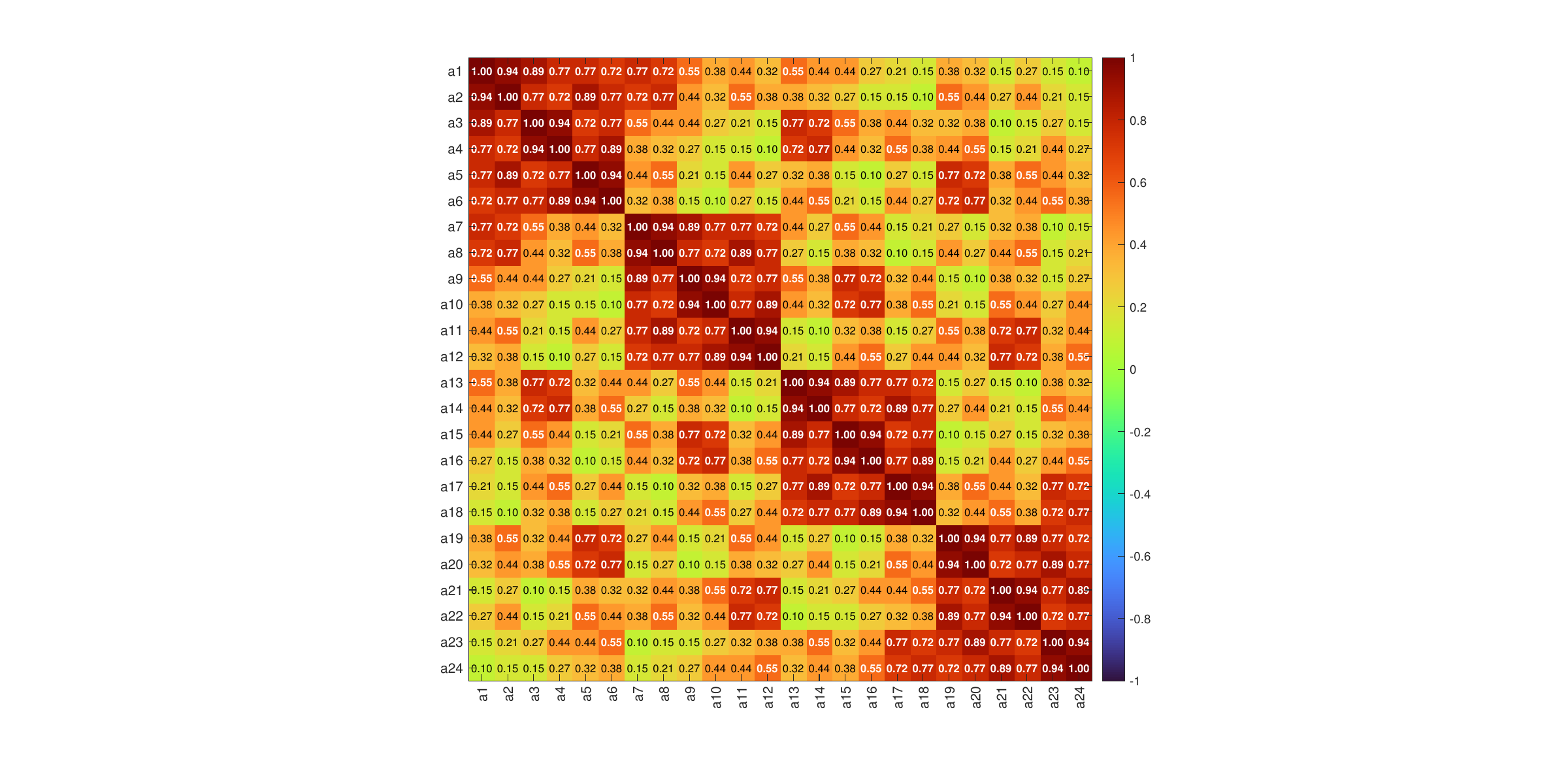}
\caption{Correlogram showing pairwise correlations between the performance profiles of the 24 permutation orders (sampling strategies) $a1$ - $a24$ (Table \ref{EVO:tab-24-compact}), i.e., variants of the same non-repeating sampling algorithm differing only in sampling order. The axes denote strategies; color intensity encodes the correlation coefficient.}
\label{fig:3}
\end{figure}

\begin{figure}[htb]
\centering
%\begin{page}{0.2\textwidth}
   %\centering
     \includegraphics[scale=0.29]{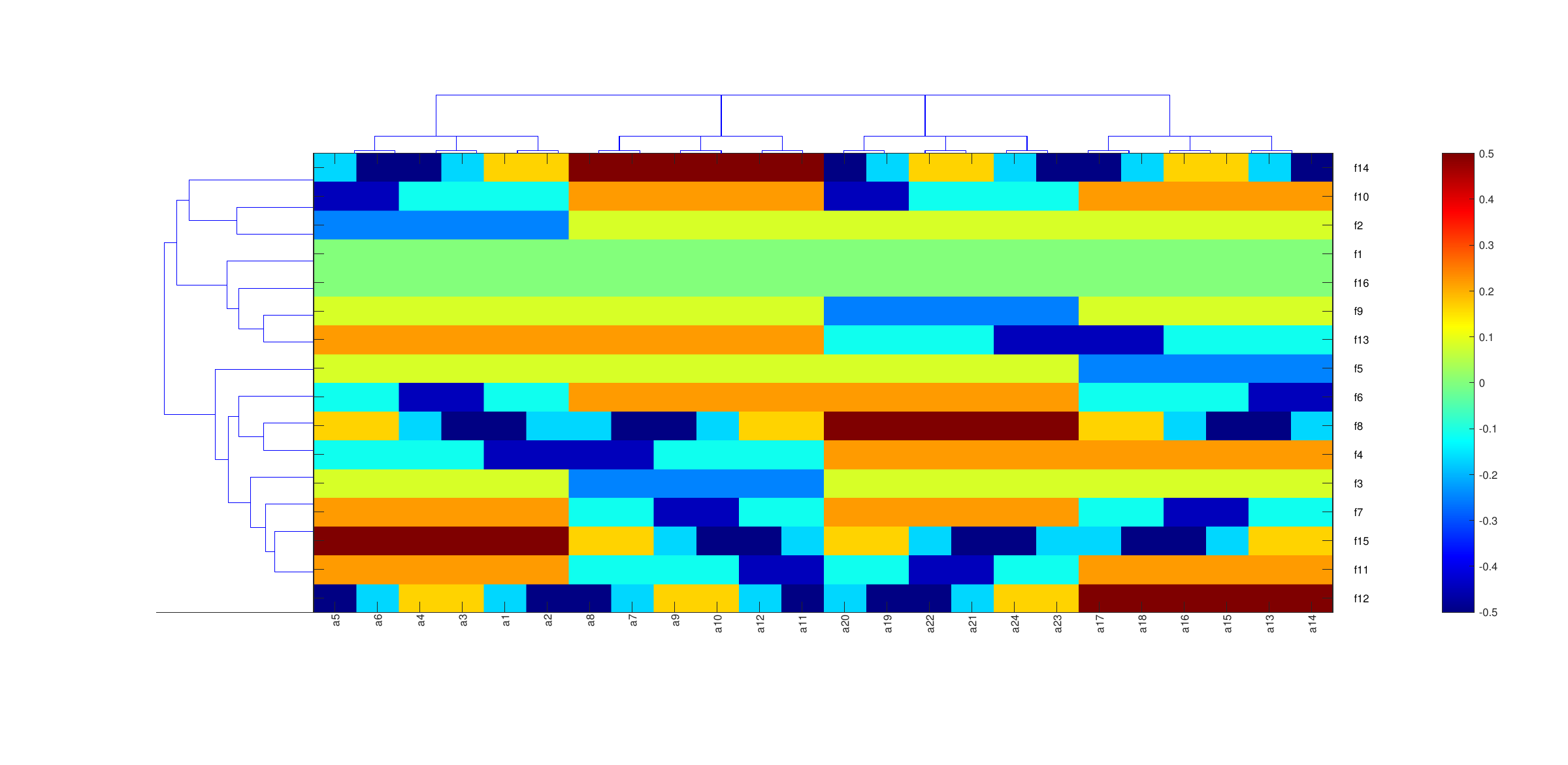}
 \caption{Hierarchical clustering of permutation based algorithms in binary function optimization.
Horizontal axis: permutation-based algorithms clustered by performance similarity. Vertical axis: Binary functions, representing algorithm behavior across different optimization landscapes.
The color scale on the right denotes row-mean-centered performance values (deviations from the function-wise mean): red and orange indicate above-average performance for a given function, while blue shades represent below-average performance. Green and yellow correspond to values close to the function-wise mean, indicating only minor deviations.}\label{fig:4}
\end{figure}

Optimization of binary functions is a fundamental problem in numerical computing and artificial intelligence. The choice of an appropriate optimization algorithm significantly impacts the efficiency of solving problems within different search spaces. This analysis is based on two datasets given in Figure \ref{fig:3} and Figure \ref{fig:4} which provide deep insights into the performance differences of various algorithms and their interactions with different test binary functions.

The correlogram in Figure \ref{fig:3} provides a detailed quantitative assessment of the relationships between 24 permutation-based algorithms given in Table 2, revealing the underlying structure of their performance across various test binary functions given in Table 1. The color-coded matrix, with correlation coefficients ranging from -1 to 1, highlights key patterns of similarity and divergence among the evaluated algorithms. Strong positive correlations, represented in shades of deep red, indicate algorithmic pairs that exhibit highly consistent performance, suggesting that their search mechanisms explore and exploit the solution space in a similar manner. Conversely, negative correlations, appearing in shades of blue, signify cases where algorithms produce fundamentally different optimization behaviors, potentially due to differing balance strategies between exploration and exploitation.

The clustering of algorithms with high correlation values suggests the presence of dominant heuristic structures that align closely in their performance. Specific algorithmic pairs $a1$ and $a2$, demonstrate near-perfect correlation, implying nearly identical search dynamics, likely due to shared heuristic design principles. Similarly, the relationship between $a3$ and $a7$ suggests that these algorithms adopt complementary strategies that respond similarly to test function landscapes, leading to stable and predictable results. The strong correlation between $a12$ and $a16$ further reinforces this trend, indicating that these algorithms maintain a balanced approach across different function sets, leading to similar convergence behaviors.

Examining the distribution of moderate correlations, certain algorithms exhibit partial alignment in their search behaviors but maintain distinct characteristics in specific function classes. The correlation between $a5$ and $a9$, for instance, suggests some degree of shared performance trends, yet noticeable variations in their efficiency on selected functions. This divergence may stem from differences in local search intensification or the ability to navigate complex solution spaces with varying constraints. Similarly, algorithms  $a10$ and $a14$ show moderate correlation values, indicating some overlap in their optimization trajectories while retaining individualized adaptation mechanisms.

The apparent distinctness of algorithms  $a19,$ $a21$ and $a23$ in the correlogram should be interpreted with caution. Although their rows in the correlation matrix differ from those of algorithms such as $a3$ and $a4$ in the original column order, sorting the correlation coefficients in ascending order yields identical profiles. This indicates that these algorithms share the same distribution (multiset) of correlation values, while differing only in the assignment of these values to specific algorithm pairs. Therefore, they should not be treated as outliers in terms of overall correlation strength. Instead, their distinction is relational and results from a different arrangement of pairwise similarities. This may affect their position in the hierarchical clustering, but it does not imply a fundamentally different global correlation profile.

The correlogram also reveals intricate structural dependencies among groups of algorithms, where clusters of strongly correlated methods emerge as functionally equivalent heuristics. These groupings indicate that some algorithms are interchangeable in terms of performance, reinforcing the hypothesis that certain heuristic designs converge toward common optimization patterns. The hierarchical nature of these relationships suggests that within a constrained function set, specific algorithms consistently outperform others, thereby challenging the assumption of uniform algorithmic effectiveness across all possible problems. This structural insight contributes to a more refined understanding of algorithmic behavior, allowing for a classification of heuristics based on their expected efficiency within controlled benchmarking environments.

The visualization of correlation structures through this correlogram serves as a critical analytical tool in evaluating algorithmic efficiency and redundancy. The identification of highly correlated algorithmic pairs offers insight into the functional overlap between heuristics, while areas of weak correlation provide an opportunity to explore complementary strategies that could enhance performance diversity. The emerging clustering patterns suggest that while certain algorithms exhibit near-identical behaviors, others diverge significantly, reflecting differences in heuristic mechanisms and adaptability to varied optimization landscapes. These observations offer a foundation for a deeper investigation into algorithmic selection strategies, particularly in designing ensembles or hybrid approaches that capitalize on complementary search behaviors.
In the literature, Surjanovic and Bingham \cite{11zxc} observed that the effectiveness of heuristic algorithms is strongly correlated with the characteristics of the optimization landscape, which is confirmed by the obtained results.

Figure \ref{fig:4} shows a hierarchically clustered heatmap of row-mean-centered performance values. It reveals structured relationships between algorithms and functions, allowing for a detailed assessment of the structural dependencies within the studied space. The visualization presents hierarchical grouping of both algorithms and functions. This enables the identification of algorithm performance patterns depending on the characteristics of the test functions. The clustering of algorithms indicates that certain methods exhibit a high level of performance consistency. This may suggest similar exploration and exploitation strategies. In particular, the group of algorithms comprising $a1, a3, a7, a10$ demonstrates stability and consistent relative performance patterns across multiple functions. This suggests that these algorithms are less susceptible to changes in the structure of test functions and can be regarded as relatively universal strategies within the analyzed benchmark.

Conversely, algorithms $a5$, $a8$, and $a12$ should not be interpreted as exhibiting a fundamentally different overall value profile based solely on the heatmap. As shown by the row-wise sorted comparisons (e.g., $a12$ vs. $a3$), the sorted profiles can be identical. In such cases, the differences arise from the assignment of values to specific functions. Thus, the distinction is function-specific (relational) rather than a difference in the overall distribution of values. This suggests function-dependent selectivity (i.e., relative specialization) rather than a globally distinct value pattern.

The clustering analysis of test functions shows that functions $f1, f2, f5$ form a homogeneous group with similar relative performance patterns. This is consistent with their similar algebraic and structural properties. It also suggests that algorithms performing well on one of these functions tend to show similarly favorable relative performance on the others within the same group. Functions $f3, f4, f6$, on the other hand, exhibit greater variability in results depending on the applied algorithm. This may imply that their structure requires specific exploration strategies. Significant performance differences are also visible for functions $f8, f9, f13$. These functions show strong local variations in row-mean-centered values. This may indicate the existence of complex function landscapes that require more adaptive optimization methods.

Particularly noteworthy are the intense contrasts in row-mean-centered values for functions $f12, f14, f15$. For some algorithms, specifically $a17, a22, a24$, these functions exhibit substantial differences. This may suggest that these algorithms employ search mechanisms that differ from those used by other methods. Additionally, the fact that certain algorithms $a9, a11, a20$ do not form distinct clusters in the hierarchical structure may indicate that their effectiveness is more heterogeneous. It may also indicate that their effectiveness depends on specific characteristics of the test binary functions.

The dendrograms also highlight closely clustered subgroups of algorithms. This is particularly visible for the set $a2, a4, a6$, which exhibits almost identical relative performance profiles across most functions. This suggests that these methods employ very similar search space exploration mechanisms. In contrast, the group $a19, a21, a23$ demonstrates much greater variability in results. This indicates that these algorithms are relatively more effective on specific subsets of test functions, but are not equally efficient across the entire problem space.

These observations are highly relevant for algorithm classification and the selection of optimization methods for specific problems. The clear clustering of functions with similar properties suggests that optimization can be effectively conducted using a set of algorithms tailored to a particular problem type. This is preferable to relying on a single method for all cases. The results also indicate that, within the studied space, there exist more challenging functions for which the differences in algorithm performance are more pronounced. This makes them a key reference point for further research on optimization methods and their adaptation to specific problem classes.

Similar observations were made by Hansen et al. \cite{6}, who propose an anytime performance assessment for black-box optimization based on the number of function evaluations needed to reach predefined targets. This framework enables quantitative performance comparisons across problem classes and evaluation budgets. This relationship is consistent with the benchmark-function descriptions compiled by Surjanovic and Bingham \cite{11zxc}. They show that complex landscapes (e.g., narrow valleys or many local minima) can make optimization difficult and can hinder convergence to the minimum.

\section{Additivity measures independent of the implementation} %\cite{10aaa}, \cite{11a},\cite{121},

Optimization algorithms play a pivotal role in addressing complex problems across a wide array of scientific and engineering disciplines. Evaluating their performance accurately and consistently is of paramount importance for both theoretical analysis and practical application. 
Traditional evaluation metrics, however, often rely on the specifics of the hardware, software environment, and implementation details. 
In black-box optimization, performance is commonly reported in objective-function evaluations rather than wall-clock time, since implementation-dependent computational overhead may dominate and can vary markedly across methods and implementations \cite{10aaa}.
This dependence creates challenges in drawing universal conclusions about algorithmic efficiency and behavior. To overcome these limitations, researchers have increasingly focused on the development of \textbf{implementation-independent measures}.  
An implementation-independent measure abstracts away hardware and software dependencies, focusing solely on the intrinsic properties of the algorithm and the objective function. Such measures are crucial for enabling a fair comparison of algorithms, irrespective of the computing environment or programming language used. They allow for the isolation of algorithmic contributions to performance and facilitate a deeper understanding of optimization principles.  In this context, one of the most intriguing properties of implementation-independent measures \cite{6},\cite{12uuu} is their \textbf{additivity}. Additivity refers to the ability to decompose a composite function into simpler components, analyze them independently, and recombine their results to predict the performance on the composite function. This property is not only theoretically elegant but also practically significant, as it offers insights into algorithm behavior and helps design more efficient strategies for complex optimization tasks.  
The question of whether an implementation-independent measure is additive is far from trivial. Additivity implies that the performance of an algorithm on a composite function $fk = fi + fj$ can be directly derived from its performance on the individual functions $fi$ and $fj$. This is of particular interest in optimization, as it allows:  \\
1. \textit{Performance Prediction}: The ability to infer performance on complex functions from simpler components reduces computational costs and accelerates the evaluation process. \\ 
2. \textit{Algorithm Design}: Understanding additivity can guide the design of algorithms that exploit the structure of composite functions for improved efficiency.\\  
3. \textit{Generalization Across Domains}: Additive measures find applications in various fields, from machine learning and game theory to risk analysis and economics, providing a unifying framework for studying diverse systems.  

Moreover, \textbf{many practical optimization problems have finite variable resolution (e.g., due to physical limits or numerical precision), which effectively discretizes the search space and motivates studying discrete and binary objective functions \cite{11a}.}
Despite its importance, the additivity of implementation-independent measures has received limited attention in the literature, particularly in the context of binary objective functions. The results presented in this paper aim to address this gap by systematically analyzing the additivity property and its implications for optimization algorithms.  
The theoretical foundation of implementation-independent measures can be traced to cooperative game theory, where the \textbf{Shapley  value} \cite{12yyy},\cite{4ah},\cite{10sz} was introduced to quantify the contributions of individual players to the total payoff of a game. 
Formally, let $v$ be a characteristic function defined on a set of features $\mathbb{N} = \{1, 2, \ldots, n\} $, where $v(S)$ represents the value of a coalition $S \subseteq \mathbb{N}$. The Shapley value $\phi_{i}(v)$,  is defined by Equation~(\ref{eq:shapley}):

\begin{equation}\label{eq:shapley}
\phi_{i}(v) = \sum_{S \subseteq \mathbb{N} \setminus \{i\}} \frac{|S|!(n - |S| - 1)!}{n!} \left[v(S \cup \{i\}) - v(S)\right]
\end{equation}
provides a fair allocation of the total value among players based on their marginal contributions. Its properties, including \textit{efficiency, symmetry, null player, and additivity}, make it a powerful tool for analyzing systems with additive characteristics.  
In the context of optimization, these properties translate to performance measures that remain invariant across different implementations and accurately reflect the algorithm’s intrinsic capabilities. Additivity, in particular, enables the decomposition of composite functions, 
offering a systematic approach to studying their behavior.  In this paper, ''decomposition'' refers to a pointwise algebraic relation between benchmark functions (e.g., $f5+f10=f14$) defined on the same domain, rather than to classical variable separability of the form $f(x,y)=g(x)+h(y)$ for all $(x, y)$ in the domain.
This study explores the additivity property of implementation-independent measures through an empirical analysis of 24 optimization algorithms applied to 16 binary objective functions. The specific objectives are:  \\
1. To establish a rigorous framework for evaluating additivity in implementation-independent measures. \\ 
2. To empirically analyze the performance of algorithms on sums and differences of binary functions and investigate the conditions under which additivity holds. \\ 
3. To identify practical implications of additivity in improving algorithm design, reducing computational costs, and understanding feature interactions. \\
The results of this study reveal several nontrivial and highly significant findings:  \\

\begin{itemize}
\item[•] \textit{Enhanced Algorithmic Efficiency}: Certain algorithms consistently perform better on composite functions, achieving optima with fewer trials. For example, algorithm $a4$ achieves the optimum for $f5 + f10 = f14$  in three trials, compared to four for $f14$  alone.  
\item[•] \textit{Systematic Patterns in Performance}: By grouping algorithms based on their numerical sequences and analyzing their behavior on composite functions, we uncover patterns that suggest a deeper structural relationship between function composition and algorithmic efficiency. 
\item[•] \textit{Broad Applicability}: The findings are not limited to binary objective functions but have potential implications for a wide range of domains, including machine learning, economics, and risk assessment \cite{12yyy},\cite{4ah},\cite{10sz},\cite{19re}.  
\end{itemize}

These results demonstrate the potential of additivity analysis to provide new insights into optimization and motivate further exploration of this property in more complex settings.  \\
This section provides a rigorous theoretical foundation for implementation-independent measures, with a particular emphasis on the property of additivity. By defining these measures and exploring their key properties, we establish a framework for understanding their role in analyzing optimization algorithms. \\ 

An \textbf{implementation-independent (performance) measure} \cite{6},\cite{12uuu} is a quantitative metric designed to evaluate the performance of optimization algorithms without being influenced by the specifics of the computational environment. Unlike conventional metrics runtime or computational complexity; these measures rely solely on the intrinsic properties of the algorithm and the objective function.  

Let $\mathcal{F} = \{f1, f2, \ldots, fn\}$ denote a set of objective functions, and let $\mathcal{A} = \{a1, a2, \ldots, am\}$  represent a set of algorithms. 
\begin{define}
Let $M(ai, fj)$ denote the number of evaluations (queries) of the objective function $fj$ required by algorithm $ai$ to discover the global minimum over the domain $\mathcal{X}$. This is equivalent to the value $s_{i,j}$ defined in Section 4.2.
\end{define}

This measure is considered implementation-independent if it satisfies the following criteria: \\ 
1. \textit{Invariance}: $M(ai, fj)$ remains unchanged under variations in hardware, software, or programming language.\\  
2. \textit{Consistency}: The measure reflects the fundamental efficiency of the algorithm in exploring the solution space. \\ 
3. \textit{Reproducibility}: Results obtained using $M(ai, fj)$ can be reproduced across different experimental setups. \\ 
Several desirable properties of implementation-independent measures ensure their utility in analyzing and comparing algorithms:  \\
(a) \textit{Efficiency}: The total number of function evaluations across all algorithms applied to a function $fj$ should match the theoretical minimum required to identify the global optimum. This property is expressed as:  $\sum_{i=1}^m M(ai, fj) \geq M^{*}(fj),$ where $M^{*}(fj)$ is the theoretical minimum number of evaluations. \\ 
(b) \textit{Symmetry}:  If two functions $fi$ and $fj$ are structurally equivalent, the measure should yield identical results for any algorithm:  $M(ak, fi) = M(ak, fj), \quad \forall ak \in \mathcal{A}.$\\
(c) \textit{Monotonicity}:  For a function $fj$ that can be decomposed into two simpler subfunctions $fj = fp + fq$, the measure should satisfy:  $M(ak, fj) \geq \max\{M(ak, fp), M(ak, fq)\}$.\\
(d) \textit{Additivity}:  One of the most critical properties, additivity ensures that the measure for a composite function can be expressed as the sum of the measures for its components. For $fj = fp + fq$, this property is formalized as:  $M(ak, fj) = M(ak, fp) + M(ak, fq).$\\ %\textit{Additivity in the Context of Binary Functions}\\  
To make the notion of additivity more precise, we formalize the operation $fk = fi + fj$ as a pointwise sum over the domain $\mathcal{X}$. That is, for every $x \in \mathcal{X}$ we define:
$fk(x) = fi(x) + fj(x)$
where $fi, fj, fk : \mathcal{X} \to \{0,1\}$ and addition is understood as integer addition. To ensure that $fk$ remains a binary function, we restrict attention to those cases where $fi(x) + fj(x) \in \{0,1\}$ for all $x$.
For example, the function $f14 = (1,0,1,1)$ arises as the pointwise sum of $f5 = (0,0,1,0)$ and $f10 = (1,0,0,1)$. However, even if $f14(x) = f5(x) + f10(x)$ holds for all $x$, the performance measure $M$ may not obey additivity. In our experiments, we observed that
$M(a4, f14) \neq M(a4, f5) + M(a4, f10)$.
This indicates that structural additivity of functions does not guarantee additive performance, which may depend on how the composition affects the landscape of the search space.\\
\textit{Adding functions may reduce symmetry in the objective landscape, thereby improving discrimination among search candidates and aiding algorithm performance.}

\section{Hierarchy of algorithms in algebraic structure-based benchmarks: new perspectives on NFL}

Modern analysis of optimization algorithms requires not only an understanding of their functioning but also a precise grasp of the structure of the problems they are tested on. In the context of the NFL theorem, an especially significant aspect is how benchmarks are organized and how they influence the performance of heuristic algorithms. One critical issue that has been overlooked in the literature is the impact of algebraic operations on test functions within sets that preserve the c.u.p. property.  
This chapter focuses on a detailed analysis of how addition and subtraction of functions affect the performance of permutation algorithms in benchmarks based on the c.u.p. structure. Specifically, we will work with 24 permutation-based algorithms listed in Table 2, evaluated on benchmarks constructed from Table 1 c.u.p. functions. These benchmarks are formed through summation and difference operations within the c.u.p. structure, based on functions from Table 1, excluding function $f1$. The resulting benchmarks retain the c.u.p. property after these algebraic transformations (The vertical axis in Figure 5 represents a c.u.p. benchmark constructed from all possible function additions within the c.u.p. set, excluding $f1$. Similarly, the vertical axis in Figure 7 represents a c.u.p. benchmark generated from all possible function subtractions within the c.u.p. set, also excluding $f1$). Such benchmarks are used in research on the robustness of heuristics to structural changes in test functions and in the construction of problem spaces with predictable optimization properties \cite{12yyy},\cite{10ak},\cite{12},\cite{28x}.
By utilizing algorithm performance correlograms, we aim to answer key research questions:  \\
\begin{itemize}
\item[•]  Do algebraic relationships between functions influence the classification of algorithms?  
\item[•]  Can addition and subtraction operations lead to local inequalities in performance despite global compliance with NFL?  
\item[•]  Do certain algorithmic strategies remain stable regardless of algebraic transformations of test binary functions?  
\end{itemize}
The addition or subtraction of functions leads to local inequalities in algorithm performance when algebraic operations modify the function landscape in a way that favors specific search strategies. 
Experimental results show that differences in algorithm efficiency arise primarily in the case of function subtraction, where some algorithms reach the optimum faster, indicating local preferences 
resulting from changes in the search space structure. While the No Free Lunch theorem remains globally valid, systematic deviations in the number of required evaluations may occur on restricted 
sets of functions, depending on how the objective function is transformed.

The analysis encompasses two fundamental aspects: \\ 

\begin{figure}[htb]
\centering
%\begin{page}{0.2\textwidth}
   %\centering
     \includegraphics[scale=0.32]{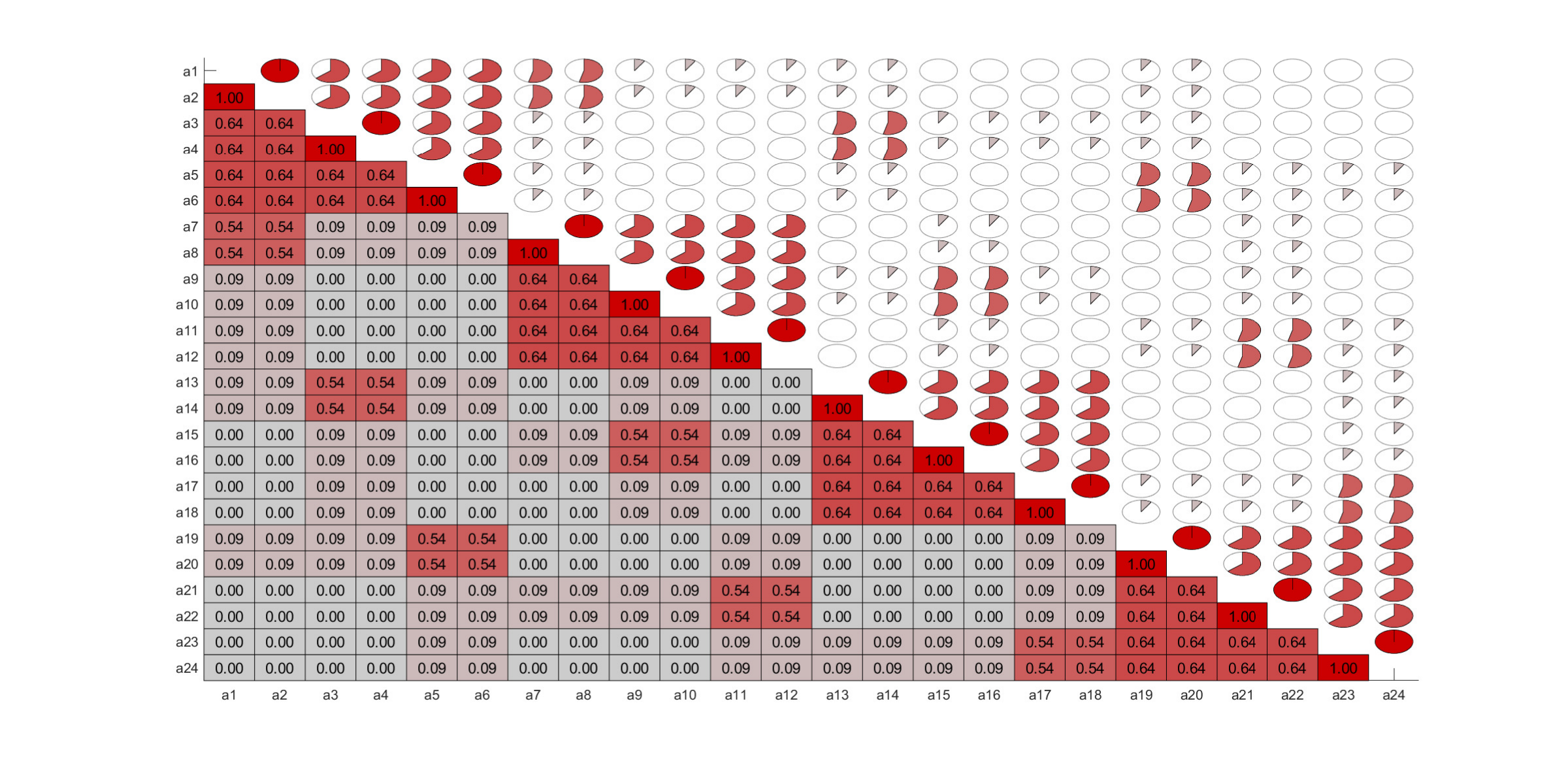}
    %\caption{\tiny{Heatmap Matrix Covariance with Clustering: Function and Algorithm.}}
%\end{page}
\caption{Correlation matrix of algorithm performance on the summation-based c.u.p. benchmark.  
Legend:  Red/Orange Shades - High positive correlation between algorithm performance.  
Blue Shades - Strong negative correlation, indicating contrasting performance trends.  
Green/Yellow Shades - Moderate correlation levels, representing partial similarity.
}
 \label{fig:5}
\end{figure}
%Figure\ref{fig:5} -correlogram and the 
\begin{figure}[htb]
\centering
%\begin{page}{0.2\textwidth}
   %\centering
     \includegraphics[scale=0.32]{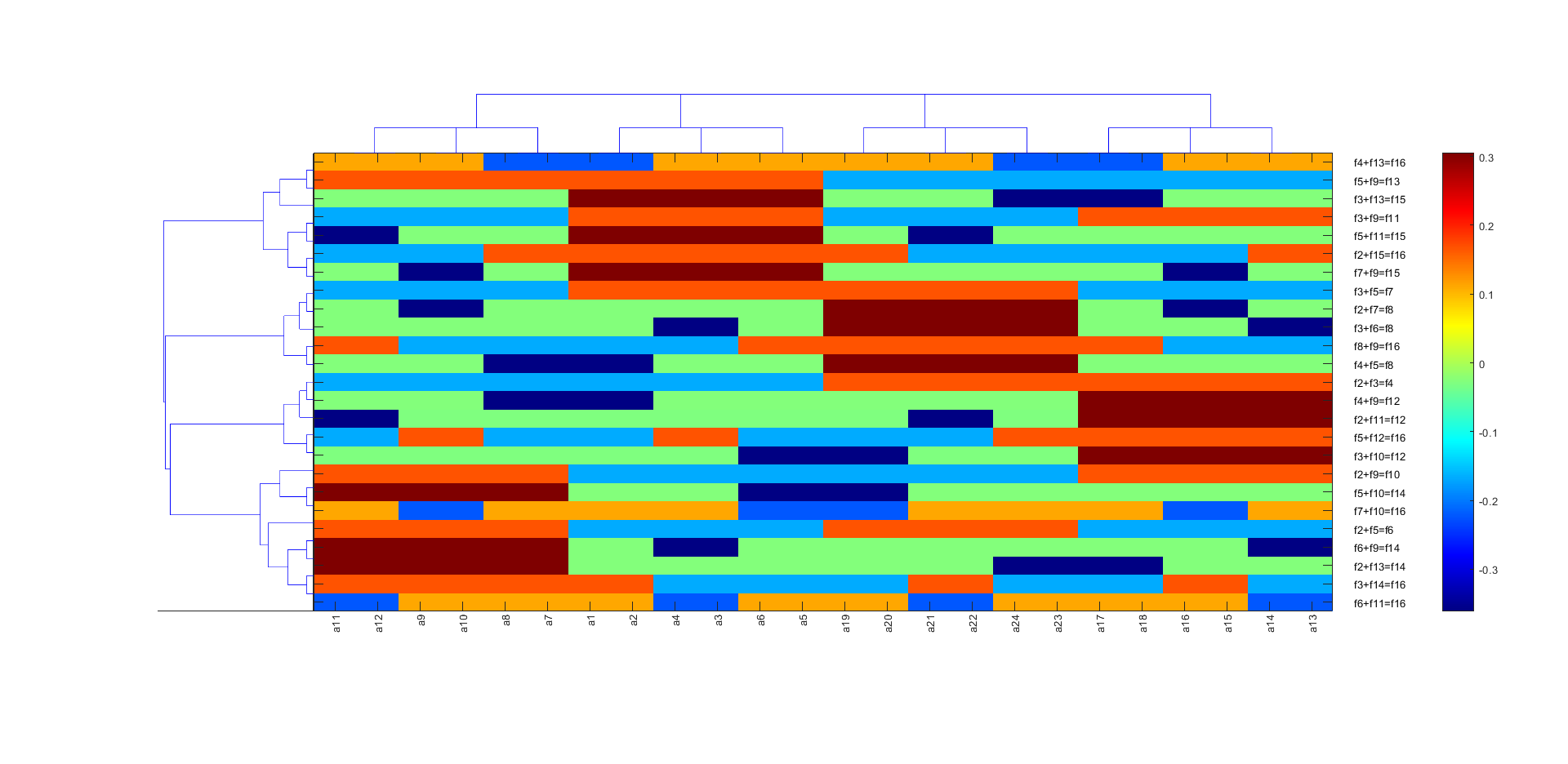}
    %\caption{\tiny{Heatmap Matrix Covariance with Clustering: Function and Algorithm.}}
%\end{page}

\caption{Hierarchical clustering of permutation-based algorithms on function additions in the c.u.p. benchmark.
Horizontal axis: permutation-based algorithms clustered by performance similarity.
Vertical axis: function additions (composite functions in the c.u.p. benchmark), representing algorithm behavior across algebraically combined optimization landscapes.
The color scale on the right denotes row-mean-centered performance values (deviations from the function-wise mean): red and orange indicate above-average performance for a given composite function, while blue shades represent below-average performance. Green and yellow correspond to values close to the function-wise mean, indicating only minor deviations.
Dendrograms show hierarchical clustering of both functions and algorithms based on similarity of performance patterns.
}\label{fig:6}
\end{figure}

\begin{figure}[h]
\centering
%\begin{page}{0.2\textwidth}
   %\centering
     \includegraphics[scale=0.32]{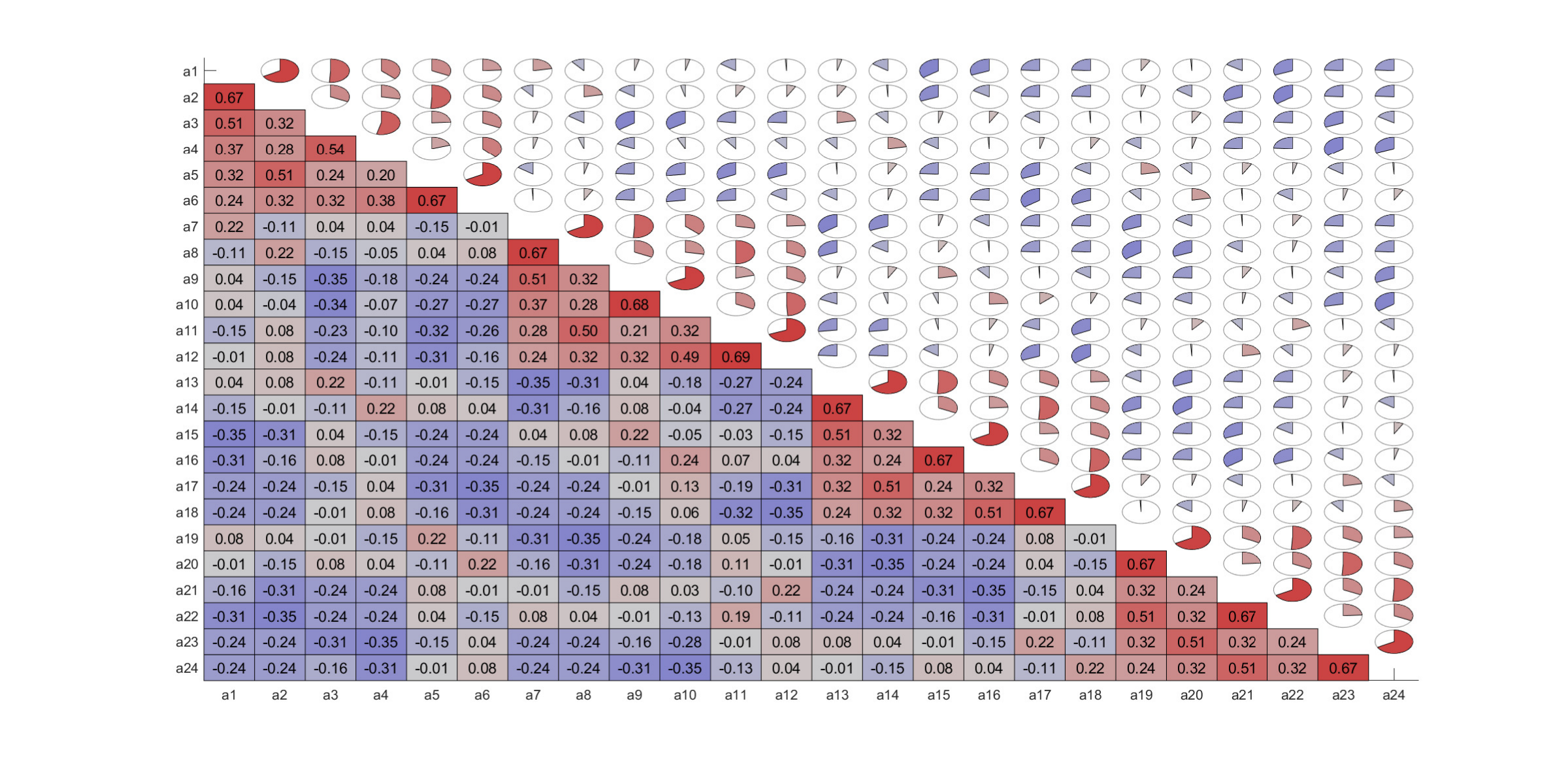}
    %\caption{\tiny{Heatmap Matrix Covariance with Clustering: Function and Algorithm.}}
%\end{page}
\caption{Correlation matrix of algorithm performance on function subtractions in the c.u.p. benchmark.  
Dark Red: high positive correlation (close to 1) - algorithms exhibit nearly identical behavior on the same function sums,  
Light Red:moderate positive correlation – algorithms show some similarity in performance trends, 
Gray: near-zero correlation - no significant relationship between algorithms on function sums,  
Circles: visual representation of correlation - the larger the red portion, the stronger the positive relationship,  
X and Y Axis: 24 permutation-based algorithms evaluated on function sums,  
Dendrogram: hierarchical clustering of algorithms based on performance similarity on function sums.}\label{fig:7}
\end{figure}
%Figure \ref{fig:7} correlogram and the 
\begin{figure}[h]
\centering
%\begin{page}{0.2\textwidth}
   %\centering
     \includegraphics[scale=0.32]{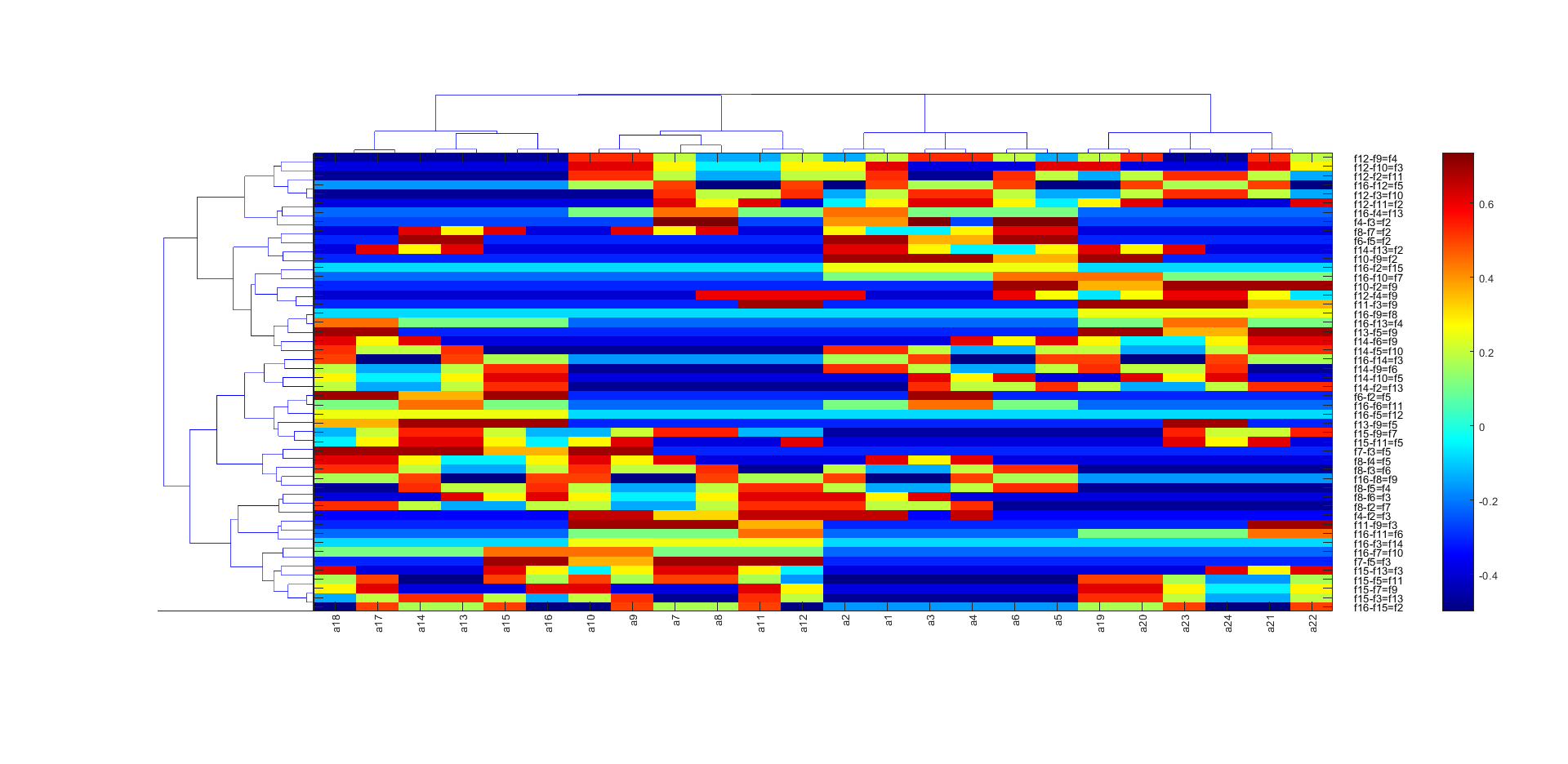}
\caption{Hierarchical clustering of permutation-based algorithms on function differences in the c.u.p. benchmark.
Horizontal axis: permutation-based algorithms clustered by performance similarity.
Vertical axis: function differences derived from the c.u.p. benchmark, representing algorithm behavior across difference-based optimization landscapes.
The color scale on the right denotes row-mean-centered performance values (deviations from the function-wise mean): red and orange indicate above-average performance for a given function difference, while blue shades represent below-average performance. Green and yellow correspond to values close to the function-wise mean, indicating only minor deviations.
Dendrograms show hierarchical clustering of both function differences and algorithms based on similarity of performance patterns. Clusters indicate groups with similar relative performance structure.}\label{fig:8}
\end{figure}

\subsection{Function Addition}
The advanced correlogram shown in Figure 4 illustrates the correlation matrix between the performances of 24 permutation-based algorithms across binary functions. 
The horizontal and vertical axes both correspond to individual algorithms labeled as  $a1$ through $a24$, representing their mutual performance correlations in the optimization process.
The matrix structure enables the identification of relationships between algorithmic behaviors, highlighting both strong and weak dependencies in performance trends.

The color gradient and circular visual elements encode the correlation coefficients between pairs of algorithms. Dark red regions indicate high positive correlations, where the performance of two algorithms is strongly related, suggesting they explore the search space in a similar manner. In contrast, near-zero correlations, represented by lighter shades, suggest independent behavior between algorithms, implying divergent search strategies. For instance, the algorithms $a1, a2, a3, a4, a5, a6$ demonstrate significant positive correlations, suggesting that these approaches share similar sampling mechanisms, leading to comparable efficiency 
profiles across tested binary functions. Similarly, another subgroup comprising  $a13, a14, a15, a16, a17, a18$ forms a separate cluster of closely correlated algorithms, reflecting their analogous performance patterns.

Conversely, algorithms such as $a9, a10, a11, a12$ exhibit weak or near-zero correlations with most other approaches, signifying that their optimization behaviors diverge substantially from those of the strongly correlated groups. This separation indicates that these methods rely on distinct exploration-exploitation balances, leading to performance variations that are not easily predictable based on the behaviors of other algorithms. The presence of isolated correlations in specific regions of the matrix further emphasizes that some algorithmic strategies are uniquely effective on certain subsets of binary functions while being inefficient in others.

The upper triangular matrix representation, utilizing circular elements, offers an additional perspective on correlation strength. The proportion of the red-filled segments in each circle visually reinforces the degree of similarity between algorithms, providing an intuitive understanding of performance dependencies. The largest fully filled circles, such as those linking $a1$ and $a2$, $a3$  and $a4$ , or $a5$ and $a6$, confirm nearly identical performance distributions, while fragmented or near-empty circles, as seen in interactions involving $a9$ or  $a12$, highlight weak interdependencies.

These findings reinforce the structural partitioning of permutation-based optimization methods into performance classes, where certain approaches exhibit consistent efficiency patterns while others follow unique search trajectories. The results suggest that function landscapes play a crucial role in dictating algorithmic behavior, and although some methods can generalize across multiple function categories, others exhibit specialized performance advantages that are difficult to extrapolate beyond specific function groups. This visualization underscores the necessity of careful algorithm selection in binary optimization tasks, as similarities in performance trends do not always guarantee universal applicability across diverse optimization landscapes.

Analyzing Figure \ref{fig:6} reveals distinct clustering and relative performance patterns for both functions and algorithms. This indicates the presence of systematic dependencies arising from algebraic operations on functions. The diagram shows row-mean-centered performance values and their hierarchical organization, which together expose structured relationships within the analyzed benchmark.

The color scheme of the diagram reflects row-mean-centered performance values for specific function sums. Red and orange areas indicate above-average performance for a given function sum (relative to the mean value of that row). Blue-shaded regions indicate below-average performance for that function sum. Green and yellow regions correspond to values close to the function-wise mean, indicating only minor deviations. For instance, functions $f4+f9=f12$, $f3+f5=f7$, and $f5+f12=f16$ exhibit notably strong local positive deviations in selected regions of the heatmap. This suggests that their algebraic compositions may create optimization landscapes with similar relative performance structure for subsets of algorithms. Algorithms $a3$, $a7$, and $a12$ demonstrate relatively consistent behavior in these regions. This may suggest a shared response pattern in the analyzed solution space.

Conversely, blue-shaded regions indicate below-average relative performance for specific function sums. An example is function $f6+f11=f16$, whose landscape appears particularly unfavorable for algorithms $a15$ and $a19$ within the analyzed benchmark. Meanwhile, functions $f2+f5=f6$ and $f2+f3=f4$ exhibit values that are closer to the function-wise mean across a broader subset of algorithms. This suggests that their influence on optimization remains more balanced and does not strongly favor any particular algorithm in relative terms.

Hierarchical clustering applied to the diagram highlights function groupings. This suggests that certain sums of test functions lead to optimization landscapes with similar relative performance patterns. Clusters of functions $f2+f9=f10$, $f3+f10=f12$, and $f5+f10=f14$ indicate shared properties that may influence algorithmic behavior in related ways. The clustering of algorithms further suggests that specific optimization methods exhibit similar performance profiles. In particular, algorithms $a7$, $a8$, and $a2$ appear to belong to the same group, which suggests a comparable response to the analyzed function sums. In contrast, algorithms $a11$ and $a22$ are positioned within more heterogeneous clusters. This suggests greater sensitivity to changes in test function structures.

A key observation derived from this analysis is that the c.u.p. benchmark structure remains organized even after the summation operation. This indicates that algebraic function combinations do not introduce complete randomness in the relative performance patterns. Instead, they preserve and reveal distinct structural dependencies. This is significant in the context of the NFL theorem, because it suggests that for specific classes of test functions within a structured benchmark, algorithms may exhibit systematic differences in performance.

Further analysis has shown that functions containing components $f2$, $f3$, $f5$, and $f9$ frequently appear in regions with pronounced local deviations from the function-wise mean. This suggests that their structural properties may play a key role in shaping the optimization landscape. Their algebraic combinations appear to influence the problem formulation in a way that benefits specific algorithms in relative terms. This observation is consistent with departures from uniformly random algorithmic efficiency within the analyzed function class.

Based on these results, critical implications arise regarding benchmark design and algorithm selection for specific optimization problems. The analysis demonstrates that instead of treating test functions as independent and random objects, their algebraic structures should be considered. These structures affect how algorithms explore and exploit the solution space. Future studies should explore this methodology in analyzing other types of algebraic operations. This would help define more precisely the dependency patterns between test binary function structures and the efficiency of optimization heuristics.

From a practical perspective, leveraging these findings for benchmark design will be crucial to better represent real-world optimization problems. Proper grouping of test functions could allow for more accurate classification of algorithms and optimization strategies tailored to the characteristics of the objective function landscape. Thus, future research could focus on developing adaptive algorithm selection methods based on the algebraic analysis of test functions. This would mark a new direction in the field of heuristic optimization.

\subsection{Function Subtraction}

The function subtraction operation introduces greater variability in results, as observed in both Figure \ref{fig:7} and Figure \ref{fig:8}.  

The presented correlogram provides a detailed analysis of the relationships among twenty-four permutation-based optimization algorithms, focusing on their covariance structure within a systematically transformed benchmark function set. The benchmark was generated by computing all possible pairwise differences between functions from the original c.u.p. set, with the exclusion of   $f1$, which consists entirely of zeros. Despite this exclusion, the newly formed function set remains closed under permutation, preserving essential structural properties that influence algorithmic behavior. The visualization of the covariance matrix reveals significant patterns in algorithm performance, highlighting both similarities and divergences in optimization strategies.

The correlogram represents algorithmic relationships through a dual-format approach, combining numerical correlation values in the lower triangular matrix with circular segment visualizations in the upper triangular section. The color scheme provides an intuitive means of assessing correlation strength, with red hues indicating positive correlations, meaning that the respective algorithms exhibit similar performance across the modified function set, while blue hues signify negative correlations, suggesting contrasting optimization tendencies. The intensity of the colors reflects the magnitude of the correlation, with stronger shades corresponding to higher absolute values. This structure enables a rapid identification of algorithm clusters that share common search behaviors, as well as pairs that tend to perform in opposition to one another.

A thorough examination of the correlation structure reveals well-defined algorithmic groupings, with certain methods exhibiting strong intra-group cohesion while maintaining distinct separation from others. Notably, the algorithms labeled as $a1$ through  $a5$ demonstrate consistently high positive correlations, indicating that these approaches rely on similar search heuristics and are affected in comparable ways by function transformations. The shared performance trends suggest deterministic tendencies, possibly driven by structured permutation strategies that remain robust across the transformed benchmark. In contrast, the algorithms $a9$ through $a12$ exhibit negative correlations with this initial group, implying fundamentally different search dynamics. Their performance divergence suggests reliance on more stochastic or adaptive mechanisms, likely incorporating probabilistic recombination or mutation techniques that allow them to explore solution spaces in a less constrained manner.

The presence of moderate correlation values in algorithms  $a15, a16$, and $a17$ suggests a more hybridized optimization strategy, blending elements of deterministic and adaptive approaches. The positioning of these algorithms within the correlation structure indicates that they benefit from function transformations in scenarios where highly structured methods encounter difficulties, yet they retain some degree of similarity with other randomized approaches. This versatility may make them particularly useful in optimization problems where maintaining a balance between exploration and exploitation is crucial.

A key observation from the correlogram is the strong positive correlation among algorithms $a19$ through $a24$, indicating a set of methods that perform in a nearly identical manner. This redundancy suggests that these algorithms may not offer unique advantages over one another, making them functionally interchangeable in many optimization contexts. The consistently strong correlations within this group imply a shared underlying search principle that responds similarly to variations in the function set, possibly reflecting a specific algorithmic bias towards certain structural properties of the transformed benchmarks.

At the same time, the correlogram highlights algorithm pairs with near-zero correlations, signifying largely independent optimization behaviors. These independent relationships are particularly important when designing ensemble approaches, as combining algorithms with uncorrelated performance can enhance overall robustness in optimization tasks. Examples of such independent behaviors can be seen in the contrasting performance profiles of algorithms $a6$ and $a20$ or $a14$ and $a22$, where their relative independence suggests that they explore the function space through fundamentally different methodologies. This insight is valuable for selecting complementary algorithms that can compensate for one another’s weaknesses, making them well-suited for hybrid optimization frameworks.

The broader implications of these findings underscore the impact of function set modifications on algorithm performance. Since the benchmark was constructed by systematically computing differences between functions within the original c.u.p. set, the resulting optimization landscapes retain inherent structural dependencies while introducing new computational challenges. The covariance structure observed in the correlogram suggests that certain algorithms exhibit greater sensitivity to these transformations, forming well-defined clusters that reflect their inherent strengths and weaknesses. The exclusion of  $f1$ ensures that trivial zero-difference cases do not distort the analysis, yet the remaining function set maintains strong interdependencies, as evidenced by the correlation patterns among the algorithms.

The analysis confirms that some permutation-based algorithms are naturally suited for optimization tasks involving function differences, while others exhibit varying degrees of adaptability depending on the transformation applied to the objective functions. This reinforces the idea that permutation-based heuristics are highly influenced by the structural modifications of their function landscapes, making their performance characteristics contingent on the properties of the underlying benchmark. Understanding these relationships allows for a more informed selection of algorithms based on covariance patterns, particularly in contexts where hybrid or ensemble approaches are being considered.

The correlogram ultimately provides a comprehensive perspective on algorithmic behavior under transformed function conditions, revealing the existence of redundant optimization strategies, oppositional search methodologies, and functionally independent techniques. The clustering of positively correlated algorithms suggests that function transformations favor certain strategies while negatively correlated methods offer alternative approaches that may be advantageous in different optimization contexts. These insights contribute to a deeper understanding of algorithmic adaptation in permutation-based optimization and offer potential directions for further research into adaptive algorithm selection frameworks based on covariance analysis.

The analysis of the hierarchically clustered heatmap in Figure \ref{fig:8} for 24 permutation-based algorithms in the context of c.u.p. function subtraction reveals distinct clustering patterns among algorithms based on their response to different function differences. The benchmark structure remains organized even after applying the subtraction operation. This suggests that certain algorithms exhibit stable behavioral patterns across subtraction-based test functions.

First and foremost, the algorithms $a1, a2, a3, a4$ form a closely clustered group, as reflected by their similar color patterns and their proximity in the clustered arrangement. Their proximity indicates that they behave in a very similar manner across many function differences. This suggests a comparable search strategy or similar heuristic mechanisms in their operation. Similarly, a strong relationship can be observed in the group $a5, a6, a7$, which also exhibits highly similar relative performance patterns. This indicates that they respond in a comparable way to the variability of subtraction-based functions.

It is also worth noting the algorithms $a10, a11, a12$, which, while forming their own subgroup, often display a contrasting relative performance pattern with respect to the first two groups. In practical terms, this means that for function differences where the $a1$ - $a4$ group tends to show above-average relative performance, the $a10$ - $a12$ subgroup may show below-average relative performance, and vice versa. This behavior may stem from a different approach to the exploration and exploitation of the solution space.

A particularly interesting case is the group $a15, a16, a17$, which, in some parts of the heatmap, exhibits similar relative performance patterns, but in other parts shows clear internal variation. This suggests that these algorithms are more sensitive to specific classes of subtraction-based functions and that their performance depends on the characteristics of the tested optimization problems. Similar irregular patterns can be observed for algorithms $a20, a21, a22$. These algorithms show visible transitions between above-average and below-average relative performance across different function differences, indicating that their operating mechanisms may respond differently depending on the applied subtraction-based function structure.

Additionally, from the color structure of the heatmap, it is evident that function differences such as $f12-f9$, $f14-f10$, and $f16-f13$ have a significant impact on the observed patterns. Their presence in intensely warm-colored regions suggests that these function differences generate strong positive deviations from the function-wise mean for specific groups of algorithms. This indicates that they are among the most influential subtraction-based functions in shaping the relative performance structure. On the other hand, function differences such as $f7-f3$ and $f11-f9$ dominate darker blue regions, indicating below-average relative performance for some algorithm groups. This suggests that algorithms respond to them in a contrasting manner, with some methods performing relatively better while others lose efficiency.

The dendrograms indicate that algorithms naturally cluster into distinct groups. This supports the conclusion that the benchmark structure remains preserved after the subtraction operation. It also allows for further classification of algorithms, not only in terms of their similarity in the classic c.u.p. setting, but also in the context of subtraction-based function differences.

Taken together, the observations from Figure \ref{fig:8} indicate that algorithms $a1$ - $a4$ and $a5$ - $a7$ exhibit strong similarities in their relative performance patterns, whereas algorithms $a10$ - $a12$ and $a15$ - $a17$ may display different behavioral patterns depending on the characteristics of the subtraction-based functions. Certain function differences have a more significant influence on the clustering of algorithms. This provides useful insight into their effectiveness in optimization analysis.

The correlation and clustering analysis allowed for the identification of groups of algorithms that exhibit robustness or sensitivity to specific test function transformations. For instance, algorithms $a3, a7$, and $a12$ display similar behavior in the context of algebraic   function subtraction, suggesting they operate according to similar exploration and exploitation patterns within the solution space. In contrast, algorithms $a15, a19$, and $a22$ exhibit distinctly different patterns depending on the analyzed functions, which may indicate their specific adaptation to certain types of objective function landscapes. 
A key observation is that the structure of c.u.p. benchmarks remains preserved even after function subtraction, demonstrating the inherent orderliness of this problem space and suggesting that, contrary to the NFL assumptions, certain algorithms may be better suited to specific function classes.

\section{The Impact of Algebraic Transformations of Test Functions on the Performance of Optimization Algorithms}

\subsection{Conclusions from Differential Heatmaps and PCA}

In this analysis, we present a detailed study of three datasets: Data1 (original data), Data2 (sum-dominant) and Data3 (difference-dominant) were obtained by algebraic recombination using sums and differences of the original functions.
Each of these datasets was analyzed using differential heatmaps and Principal Component Analysis (PCA) to investigate the impact of algebraic operations on the relationships between functions and algorithms. The goal of this analysis is to understand how algebraic operations on functions affect the structure 
of the data and the behavior of algorithms in the context of the No Free Lunch Theorem.
Data1 represents the original functions from Table 1, which belong to the c.u.p.  set. This set includes functions from $f2$ to $f15$, excluding $f1$ (composed entirely of zeros) and $f16$ (composed entirely of ones). The performance of these functions is 0 and 1 for each algorithm, respectively, due to their definitions. Data1 
thus contains a matrix of performance values for 14 functions $f2-f15$ and 24 algorithms $a1-a24$. Each function in Data1 is independent and represents the original relationships between functions and algorithms.
Data2 was created by replacing the original functions from Table 1 with sums and differences of functions from the same c.u.p. set. Each function in Data2 is the result of algebraic operations on functions from Data1. For example:\\
- $f2$ in Data2 is the result of the operation $f4 - f3$,\\
- $f3$ in Data2 is the result of the operation $f8 - f6$,\\
- $f4$ in Data2 is the result of the operation $f12 - f9$,\\
- $f5$ in Data2 is the result of the operation $f14 - f10$,\\
- $f6$ in Data2 is the result of the operation $f2 + f5$,\\
- $f7$ in Data2 is the result of the operation $f3 + f5$,\\
- $f8$ in Data2 is the result of the operation $f3 + f6$,\\
- $f9$ in Data2 is the result of the operation $f13 - f5$,\\
- $f10$ in Data2 is the result of the operation $f2 + f9$,\\
- $f11$ in Data2 is the result of the operation $f3 + f9$,\\
- $f12$ in Data2 is the result of the operation $f3 + f10$,\\
- $f13$ in Data2 is the result of the operation $f5 + f9$,\\
- $f14$ in Data2 is the result of the operation $f2 + f13$,\\
- $f15$ in Data2 is the result of the operation $f7 + f9$.\\

These operations introduce new relationships between functions and algorithms, which may lead to changes in algorithm performance.

Data3 was created by replacing the original functions from Table 1 with differences and sums of functions from the same c.u.p. set. Each function in Data3 is the result of subtracting two  functions from Data1. For example:\\
- $f2$ in Data3 is the result of the operation $f6 - f5$,\\
- $f3$ in Data3 is the result of the operation $f4 - f2$,\\
- $f4$ in Data3 is the result of the operation $f8 - f5$,\\
- $f5$ in Data3 is the result of the operation $f7 - f3$,\\
- $f6$ in Data3 is the result of the operation $f8 - f3$,\\
- $f7$ in Data3 is the result of the operation $f15 - f9$,\\
- $f8$ in Data3 is the result of the operation $f4 + f5$,\\
- $f9$ in Data3 is the result of the operation $f15 - f7$,\\
- $f10$ in Data3 is the result of the operation $f14 - f5$,\\
- $f11$ in Data3 is the result of the operation $f15 - f5$,\\
- $f12$ in Data3 is the result of the operation $f11 + f2$,\\
- $f13$ in Data3 is the result of the operation $f15 - f3$,\\
- $f14$ in Data3 is the result of the operation $f9 + f6$,\\
- $f15$ in Data3 is the result of the operation $f13 + f3.$\\

These operations introduce another layer of changes in the relationships between functions and algorithms.

\begin{figure}[htb]
\centering
%\begin{page}{0.2\textwidth}
   %\centering
     \includegraphics[scale=0.32]{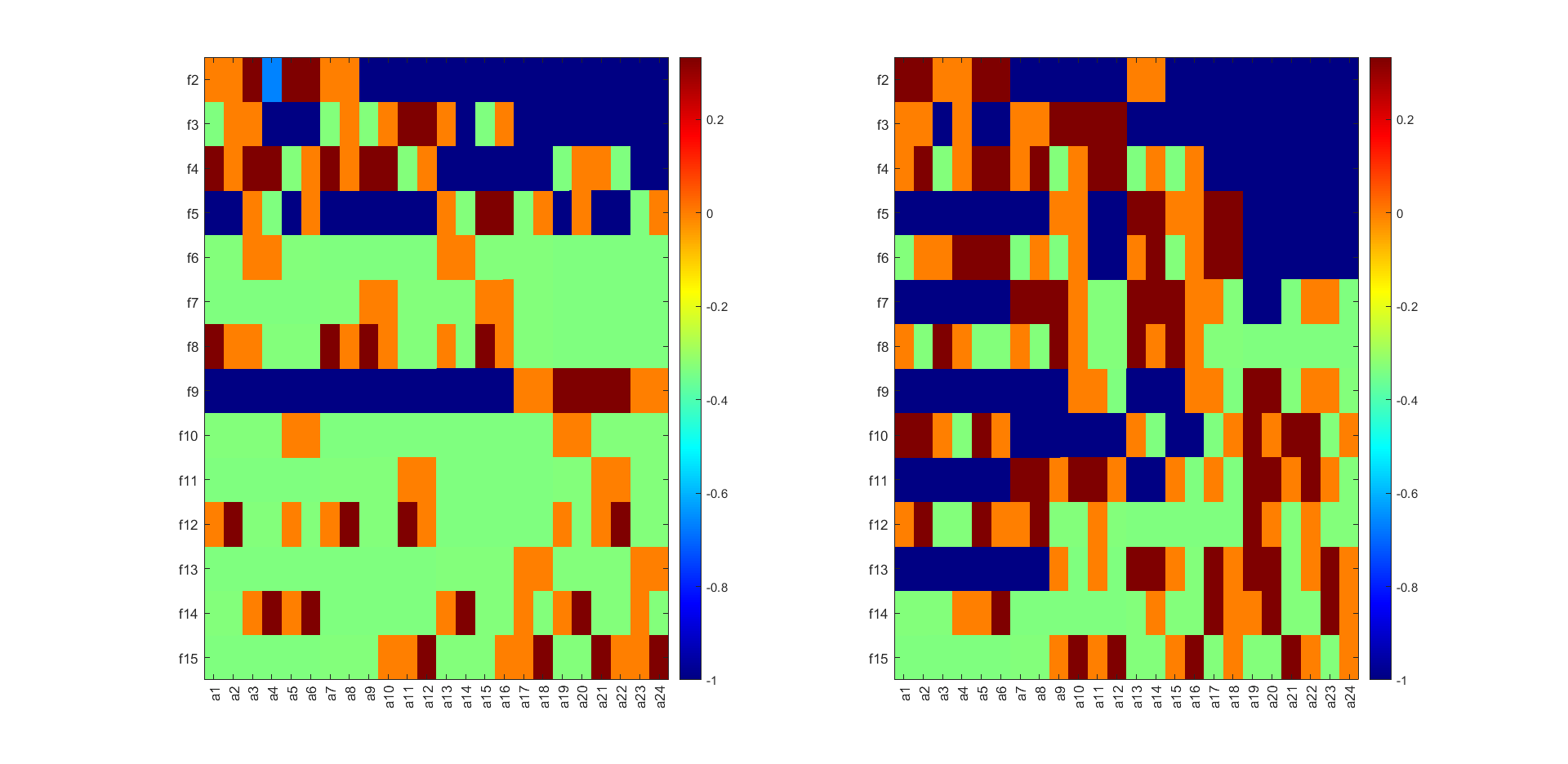}
    %\caption{\tiny{Heatmap Matrix Covariance with Clustering: Function and Algorithm.}}
%\end{page}
\caption{Comparison of function-algorithm relationships with modified functions. a) Delta Heatmap: Data2 - Data1, b) Delta Heatmap: Data3 - Data1}
 \label{fig:9}
\end{figure}
%Figure\ref{fig:5} -correlogram and the 
\begin{figure}[htb]
\centering
%\begin{page}{0.2\textwidth}
   %\centering
     \includegraphics[scale=0.32]{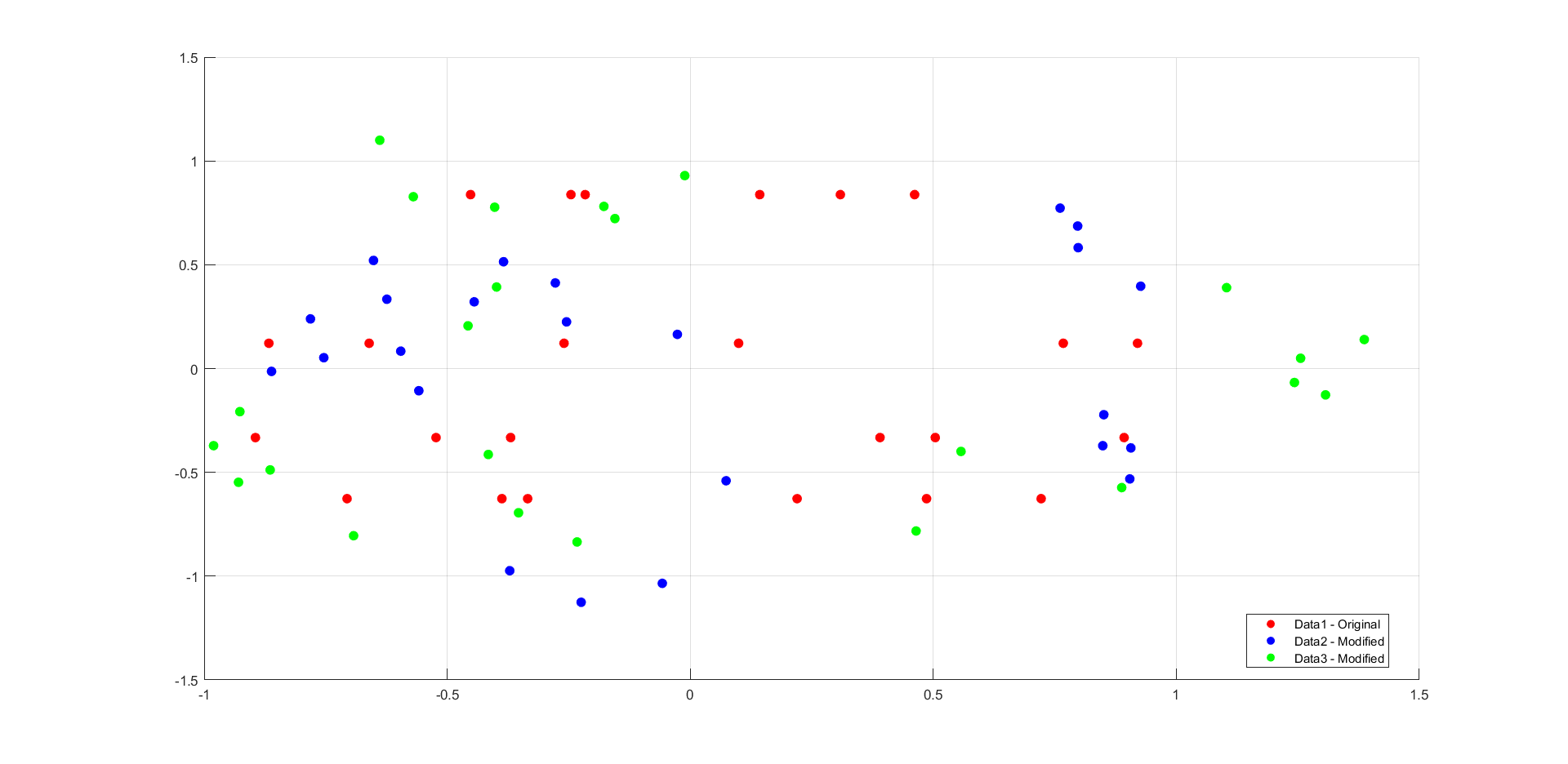}
    %\caption{\tiny{Heatmap Matrix Covariance with Clustering: Function and Algorithm.}}
%\end{page}
\caption{PCA for Three Data Sets.  X-axis: First Principal Component (PC1), Y-axis: Second Principal Component (PC2)}\label{fig:10}
\end{figure}

Figure \ref{fig:9} compares function-algorithm relationships after algebraic modification of the benchmark functions using two delta heatmaps. Panel (a) reports the element-wise differences Data2-Data1. Panel (b) shows 
Data3-Data1. The colormap encodes both the sign and the magnitude of the change in performance values. Red indicates positive deviations (higher values in the modified dataset). Dark blue indicates strong negative deviations (lower values). Orange/yellow denotes small deviations around zero. Green/cyan corresponds to moderate negative shifts.

In panel (a), negative deviations dominate and form coherent, block-structured regions. The most pronounced degradation is observed for $f2.$ An extended deep-blue region spans approximately $a9 - a24.$ This indicates a large decrease relative to the baseline across many algorithms. A second major negative structure occurs for $f9.$ It is nearly uniformly deep blue over $a1 - a16.$ This row then exhibits a clear sign reversal for later algorithms. Around $a19 - a22,$ differences become strongly positive (red). This shows that the same algebraic construction can penalize broad algorithm groups while benefiting a narrower subset. 
Outside these large-scale blocks, many entries take green/cyan tones. This is consistent with a systematic moderate decrease rather than isolated fluctuations. Additional severe local drops are also visible (e.g., $f5 - a21$ and $f5 - a22$). Positive deviations are present but remain localized. These include $f8 - a1$ and $f4-a9/a10.$ They also include several gains for $f12$ (e.g., $a2,$ $a11,$ $a22$) and multiple gains for $f14$ (e.g., $a4,$ $a6,$ $a14,$ $a20$). Near-zero changes (orange/yellow) are comparatively sparse. They appear mainly as isolated cells.

Panel (b) displays a more heterogeneous, higher-contrast pattern. There is frequent alternation between strongly negative and strongly positive regions. Large negative areas persist. This is most notable for $f2$ across approximately $a7 - a12$ and again $a15 - a24.$ It is also visible for $f5$ over early algorithms $a1 - a8.$ At the same time, panel (b) contains larger and more contiguous positive blocks than panel (a). This indicates stronger pair-dependent re-ranking. Clear examples include $f3$ over $a9 - a12.$ They also include $f5$ around $a13/a14$ and $a17/a18.$ Another example is $f7$ around $a7 - a9$ and $a13 - a15.$ Additional positive clusters appear for higher-index functions with selected later algorithms. For instance, $f10$ shows positives around $a19/a21/a22.$ Similarly, $f13$ shows positives around $a13/a14/a17/a19/a20/a23.$ Moderate negative background regions (green/cyan) remain visible for parts of $f8$ and $f14.$ Because Data3 is constructed primarily from differences but also includes several sums, the pattern in panel (b) should be interpreted as the net effect of this algebraic recombination. It should not be interpreted as a purely subtractive perturbation.

Figure \ref{fig:10} reports a PCA of three datasets: Data1 (red), Data2 (blue), and Data3 (green). PCA is computed on transposed matrices Data1, Data2, Data3. Each point is one algorithm $a1 - a24.$ Each algorithm is represented by a 14-dimensional profile over $f2 - f15.$ The plot shows PC1 vs. PC2 scores. 
PCA is performed separately for each dataset. Thus, the axes are dataset-specific orthogonal bases. Interpretation follows explained variance, dominant loadings, and within-dataset score structure.

For Data1, PC1 and PC2 explain $30.22\%$ and  $30.22\%$ of variance ($60.45\%$ total). A defining feature is the PC1–PC3 degeneracy. Each of PC1-PC3 explains $30.22\%.$ The leading variability lies in a 3D subspace. The specific axes inside that subspace are not unique. We therefore use aggregated squared-loading contributions over PC1-PC3. The largest contributions come from $f8,$ $f12,$ $f14,$ $f15.$ Each contributes 0.439 over PC1-PC3. A second tier is $f4,$ $f6,$ $f7,$ $f10,$ $f11,$ $f13.$ Each contributes 0.173. The smallest contributions are $f2,$ $f3,$ $f5,$ $f9.$ Each contributes 0.052. Data1 is strongly low-dimensional. Its dominant variability is split across three equally important directions.

For Data2, PC1 explains $33.99\%$ and PC2 explains $23.52\%$  ($57.51\%$  total). PC1-PC3 explain $74.87\%.$ Loadings show the dominant contrasts induced by sum/difference recombination. Along PC1, the largest absolute loadings are $f9$ (0.628) versus $f3$ (0.501) and $f2$ (0.390). Signs are conventional. The primary separation is therefore driven by $f9$ against the joint contribution of $f3$ and $f2.$ Along PC2, the leading contrast is $f4$ (0.572) versus $f5$ (0.480). Additional non-negligible terms are $f12$ (0.312) and $f3$ (0.273). Score extremes mirror these contrasts. Along PC1, the most extreme algorithms  include $a20$ (0.927), $a24$ (0.906), $a18$ (0.904), and $a3$  (0.861). Along PC2, the largest magnitudes occur at $a16$ (1.126), $a15$ (1.035), and $a13$ (0.974).

For Data3, PC1 explains $32.45\%$ and PC2 explains $19.01\%$ ($51.46\%$ total). PC1-PC3 explain $68.31\%.$ Data3 also has the largest overall variability across algorithms. Total variance (sum of eigenvalues) increases from 1.053 (Data1) to 1.264 (Data2) and reaches 1.992 (Data3). Loadings identify the dominant drivers. Along PC1, $f2$ (0.507)  contrasts with $f11$ (0.443), $f9$ (0.408), and $f13$ (0.325). Along PC2, $f3$ (0.580) contrasts with $f10$ (0.529). Further contributions come from $f4$ (0.360)  and $f6$ (0.261). Score extremes are stronger than in Data2. For PC1, the largest magnitudes are $a2$ (1.387), $a5$ (1.307), $a6$ (1.256), and $a20$ (0.981). For PC2, the strongest magnitudes include $a10$ (1.100), $a8$ (0.929), and $a21$ (0.835).

Taken together, Figures \ref{fig:9} and \ref{fig:10} show that algebraic recombination reshapes function-algorithm relations in a structured, non-uniform manner. In Figure \ref{fig:9}a), Data2-Data1 is dominated by broad, block-structured reductions, with fewer localized gains and occasional sign reversals for specific algorithm subsets. In  Figure \ref{fig:9}b), Data3-Data1 is higher-contrast and more heterogeneous, with substantial improvements and degradations concentrated in distinct subregions, consistent with stronger pair-dependent re-ranking. Figure  \ref{fig:10} supports this at the global level: Data2 is driven primarily by the 
$f9$ vs. ($f3$, $f2$) contrast, with a secondary  $f4$ - $f5$ contrast, whereas Data3 is dominated by $f2$ vs. ($f11,$ $f9,$ $f13$) and $f3$ vs. f$10.$ Overall, the transformations selectively alter dominant function-level contrasts, producing dataset-specific reorganization of algorithm behavior.
These transformations systematically reshape the dominant function-level contrasts, yielding dataset-specific reordering of algorithm performance profiles.

\subsection{Analysis of Study Results Using ANOVA and Tukey's Post Hoc Test}

This analysis aimed to determine whether modifications to benchmark functions significantly affect optimization algorithm performance, even within the c.u.p. function space.
A one-way analysis of variance (ANOVA) \cite{1010} was employed to assess the differences between the datasets. The $F$-statistic and corresponding $p$-value were computed to test the null hypothesis, which posits no significant differences between the mean results in the respective datasets. 
In addition, Tukey’s post hoc test \cite{1111} was conducted to investigate detailed pairwise differences between datasets.
The results of the ANOVA are presented in Table 5. The calculated $F$-statistic value is 110.68, and the associated $p$-value is  $3.59965 \times 10^{-44},$ which is significantly below the commonly used threshold ($\alpha = 0.05$). This leads to the rejection of the null hypothesis, indicating that 
the differences between the three datasets are statistically significant.

\begin{table}[h!]
\tiny{
\centering
\begin{tabular}{|c|c|c|c|c|c|}
\hline
Source of Variance & Sum of Squares (SS) & Degrees of Freedom (DF) & Mean Square (MS) & F - statistic & p-value \\ \hline
Data             & 24.533                     & 2                               & 12.2666                   & 110.68            & $3.59965 \times 10^{-44}$ \\ \hline
Error            & 111.379                    & 1005                            & 0.1108                    & -                   & -                \\ \hline
Total            & 135.913                    & 1007                            & -                             & -                   & -                \\ \hline                                             
\end{tabular}}
\caption{Summary of ANOVA Results} %\Table 5.
\end{table}

Table 5 reports a highly significant main effect of the dataset in a one-way ANOVA, indicating that at least one dataset differs in its mean element-wise performance value. Because the ANOVA F-test does not identify which specific pairs of datasets differ, Tukey’s  post hoc procedure was applied to all pairwise 
contrasts among Data1, Data2, and Data3. The resulting mean differences, confidence intervals, and adjusted p-values are summarized in Table 6.

\begin{table}[h!]
\tiny{
\centering
\begin{tabular}{|c|c|c|c|c|}
\hline
Comparison         & Mean Difference & Lower Confidence Interval & Upper Confidence Interval & p-value \\ \hline
Data1 vs. Data2    & 0.32639                 & 0.26619                            & 0.38658                             & 0.0000           \\ \hline
Data1 vs. Data3    & 0.33532                   & 0.27512                            & 0.39551                             & 0.0000           \\ \hline
Data2 vs. Data3    & 0.00893                  & -0.05127                            & 0.06912                            & 0.93556           \\ \hline         
\end{tabular}}
\caption{Results of Tukey's Test} %\Table 6.
\end{table}

\textbf{Interpretation of Tukey's Test Results}\\
1. Data1 vs. Data2: The mean difference was 0.32639, and the $p-$value = 0.0000, indicating a statistically significant difference between these datasets.\\
2. Data1 vs. Data3: The mean difference was 0.33532, and the $p-$value = 0.0000, also confirming a significant difference.\\
3. Data2 vs. Data3: The mean difference was 0.00893 and the $p-$value = 0.93556, suggesting no significant differences between these datasets.\\

Modifications of functions through addition and subtraction alter the structure of optimization problems, affecting algorithm efficiency even though the functions remain within the c.u.p. set. PCA analysis \cite{1616} and ANOVA tests indicate that these operations introduce statistically significant shifts in performance 
relative to the original dataset, leading to variations in algorithm behavior. Algorithms based on more exploratory strategies appear to be less affected by these changes compared to those that heavily rely on local function structures.

\subsection{Analysis of the Boxplot for Data1, Data2, and Data3}

\begin{figure}[htb]
\centering
%\begin{page}{0.2\textwidth}
   %\centering
     \includegraphics[scale=0.27]{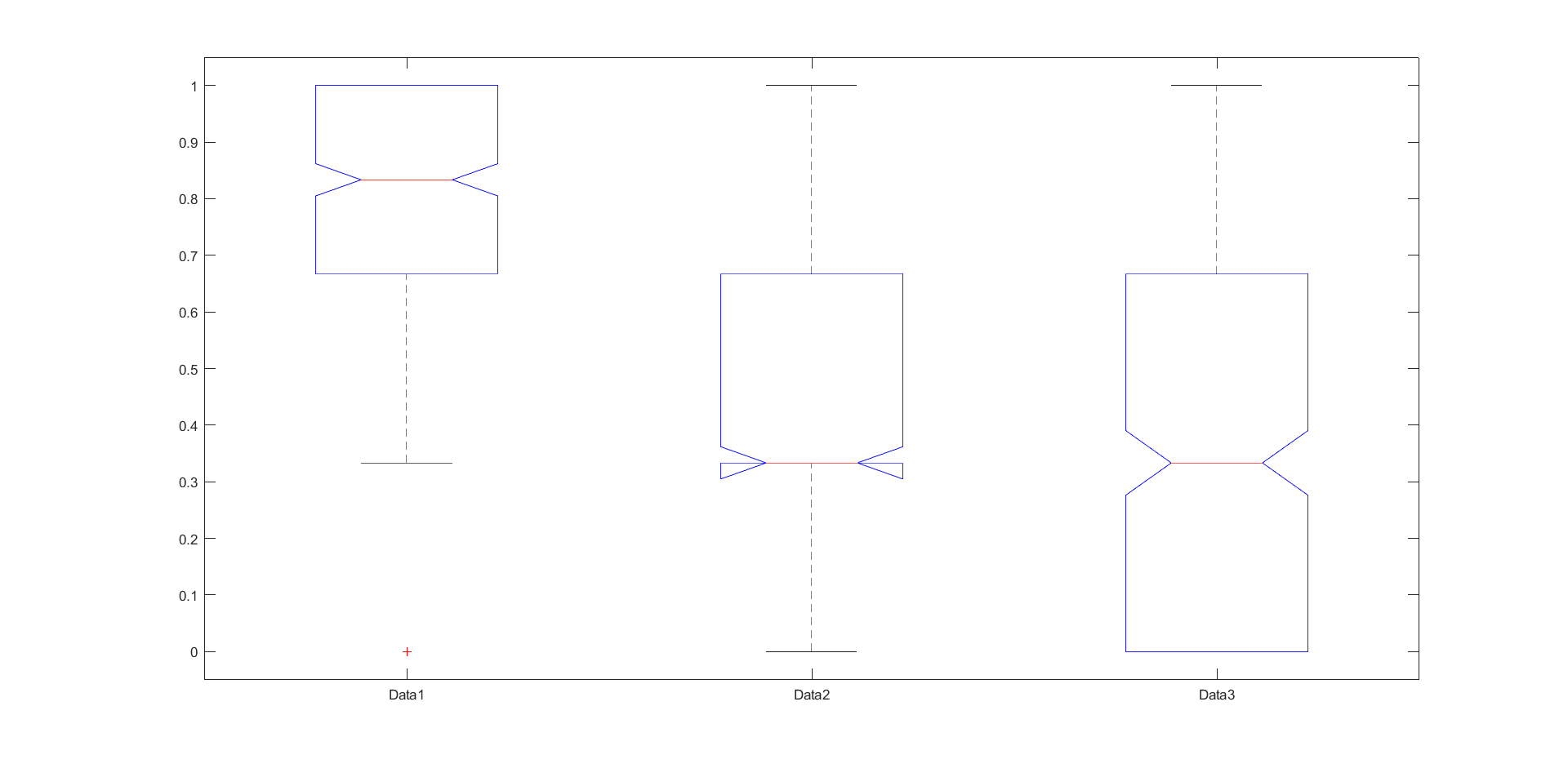}
    %\caption{\tiny{Heatmap Matrix Covariance with Clustering: Function and Algorithm.}}
%\end{page}
\caption{Comparison of Distributions of Three Data Sets Using a Box Plot. X-axis: Data Sets, Y-axis: Function Values}\label{fig:11}
\end{figure}

The boxplot in Figure \ref{fig:11} illustrates the distribution of values for three datasets: Data1, Data2, and Data3. Each of these datasets contains results obtained through different algebraic transformations of test functions, allowing for an assessment of their impact on the performance of optimization algorithms. The boxes in the plot represent the interquartile range (IQR), which spans the values between the first (Q1) and third quartiles (Q3). The red lines within the boxes indicate the median, while the whiskers show the range of values excluding outliers. Individual points outside the whiskers represent outliers, which exceed 1.5 times the IQR.

Analyzing Data1, we observe that the values in this dataset are highly concentrated around high values, indicating stability and uniformity in the results concerning the tested functions. The interquartile range oscillates around 0.8 - 0.9, while the lower bound is notably restricted. Outliers appear in the lower part, suggesting that certain functions and algorithms exhibited significantly poorer performance in rare instances.
For Data2, a much greater dispersion of values is evident compared to  Data1. The central value (median) has shifted towards significantly lower values, indicating an overall decrease in algorithm performance within this dataset. The box is asymmetric, with an extended lower part, suggesting that results in the lower range are more scattered. The whiskers extend to values close to zero, indicating that some test functions result in substantial performance degradation for algorithms.
Data3 exhibits a distribution similar to  Data2, although its values appear even more dispersed. The results span nearly the entire range from values close to zero up to 1.0, meaning that certain algorithms achieve very high efficiency on selected functions, while others display extremely low effectiveness. Additionally, the box in this case is larger than in Data2, suggesting even greater variance in results and a stronger differentiation in algorithm behavior. The interquartile range encompasses both very low and high values, meaning that the structure of test functions, after undergoing transformations, leads to extreme results for different optimization methods.
Comparing the three datasets, clear differences emerge as a result of the applied modifications to the test functions. Data1, as the original dataset, exhibits the highest stability and relatively uniform values, suggesting that the original set of test functions was well-balanced in relation to the algorithms. Data2 and Data3, in which functions were subjected to summation and differentiation operations, show significantly greater variance, as well as a noticeable downward shift in the median. The fact that values in these datasets span the full spectrum from 0 to 1 indicates complex interactions between algorithms and transformed test functions, ultimately causing optimization algorithms to exhibit non-uniform performance.
The results of the ANOVA and Tukey's test indicate that modifications to test functions have a significant impact on algorithm results. The original dataset (Data1) significantly differs from both modified datasets (Data2 and Data3), while the differences between Data2 and Data3 are statistically insignificant. 
This suggests that both types of modifications to test functions lead to similar changes in algorithm performance.
In the context of the NFL theorem, these results have significant implications. Even in permutation-closed spaces, modifications to test functions can lead to situations where the NFL theorem does not hold. This means that the choice of test functions is crucial for the evaluation of optimization algorithms.

\vspace{0.11cm}

The following theorem is supported by our empirical findings:

\begin{tw}
The No Free Lunch Theorem (NFLT) holds globally in function spaces that are closed under permutation (c.u.p.) and are sampled uniformly, as formally proven in Igel and Toussaint \cite{9},\cite{91}.  
However, introducing algebraic modifications (e.g., addition or subtraction) to benchmark functions can induce structural biases that violate uniformity assumptions, potentially leading to local  violations of NFLT - even within a c.u.p. space.
\end{tw}

\begin{corol}
NFL does not hold locally in test function sets modified through algebraic operations, even if the original structure was c.u.p.
\end{corol}

In general, the fact that the NFL does not hold on subsets (apart our paper, similar sets were considered, among others, in \cite{91}) of the set of all functions seems natural, but nonetheless, the rigorous mathematical statements in this area are interesting.
 The NFL holds on a subset if and only if it is a c.u.p. closure \cite{3},\cite{91}. \\

If this is the case, the NFL property can be considered independently from the set structure, as formulated in the following:

\begin{obs}
A c.u.p. closure transformed in this way is not a c.u.p closure anymore.
\end{obs}

\begin{implication}
Given Observation 1, it follows that NFL does not hold for the transformed data set analyzed in this work.
\end{implication}

\subsection{Monte Carlo Simulations for Higher Dimensions: Methodology, Results, and Comparative Analysis}

To generalize the conclusions drawn from the exhaustive analysis conducted for the case of $n=4$, an extensive empirical study was carried out using the Monte Carlo method \cite{4a} for larger problem sizes: $n=10$, $n=30$, $n=50$, and $n=100$. The objective of these experiments was to determine whether the structural relationship between the permutation-based sampling order and the optimization performance of objective functions persists in exponentially more complex spaces, where full enumeration becomes computationally infeasible.

In accordance with the assumptions and positions established earlier in the study, representative sampling of binary functions $f:\mathcal{X} \to \{0,1\}$ with specific structural characteristics was applied for larger values of $n$. Three function classes were considered: balanced functions, where the number of zeros and ones is approximately equal; strongly unbalanced functions, in which at least 91\% of output values are identical; and symmetric functions, whose outputs depend solely on the Hamming weight of the input. For $n=10$ and $n=30$, 50 functions were analyzed (approximately 16 - 17 from each class); for $n=50$, 15 functions were studied (5 per class), and for $n=100$, three functions were examined-one from each class. Given the enormous complexity of the function space, which grows double-exponentially with the size of the domain (for $n=100$, it reaches $2^{2^{100}}$), such a selection represents a compromise between representativeness and computational feasibility.

The Monte Carlo method was implemented following methodological guidelines from the literature \cite{4xx}, \cite{4a}, ensuring control over estimator variance and representativeness of the sample in the context of large function spaces.

In each experiment, non-repeating sampling with a budget of $m=4$ points was employed, following the methodology used in the $n=4$ case. For $n=10$, all 24 permutations were tested; for $n=30$, 10 representative permutations were selected; for $n=50$, 6 were used; and for $n=100$, 3 algorithms were applied. For each function, the best result found was recorded, and the average efficiency of each algorithm was computed.

For $n=10$, differences between algorithms were clearly observable. For balanced functions, the difference in performance (mean best result) between the best and worst algorithm exceeded 18\%. For symmetric functions, some permutations produced results that were 12\%-15\% better than others. Strongly unbalanced functions showed much smaller variance-nearly all permutations yielded near-optimal results, ranging from 35.4\% to 62.1\%, with an average around 49.2\%. For balanced functions, the results ranged from 46.7\% to 78.2\%, and for symmetric functions-from 58.9\% to 81.3\%, with a clear dominance of exploratory permutations.

For $n=30$, analogous patterns were observed: the performance gap between the best and worst-performing permutations reached 22 - 30 percentage points. The average efficiency for balanced functions was 63.4\% (range: 52.3\% - 75.5\%), for unbalanced functions 49.5\% (40.1\% - 58.8\%), and for symmetric functions 70.6\% (60.2\% - 84.0\%). High performance of exploratory permutations remained evident in this setting as well.

For $n=50$, due to the limited sample size (15 functions, 6 algorithms), the analysis served as a more focused exploratory study. The results still revealed considerable variance. For symmetric functions, the difference between the best and worst permutation reached 21 percentage points; for balanced functions, 15 points; and for unbalanced functions, about 6 - 8 points. The average efficiencies were 66.1\%, 59.2\%, and 46.5\%, respectively-consistent with the trends observed for lower dimensions.

In the case of $n=100$, one function from each class was analyzed, each tested with three different algorithms. The differences between permutations for the symmetric function reached 19.4\%, for the balanced function 12.6\%, and for the unbalanced function 6.3\%. Despite the small sample size, the observed trends remained consistent, and the structure of the sampling continued to significantly affect the final result.

The conducted study confirms that the relationship between the sampling order and algorithm performance is not merely an artifact of the limited space in the $n=4$ case but rather a structural phenomenon that persists in higher-dimensional settings. Algorithmic efficiency depends on both the function class and the type of permutation used. These results challenge the universal validity of the classical No Free Lunch theorem and suggest that, in practical optimization scenarios, the structure of the algorithm and the nature of the objective function play a critical role. Symmetric and balanced functions are particularly sensitive to the sampling strategy, which opens possibilities for more precise characterization and the design of dedicated optimization strategies. The Monte Carlo method, supported by the selection of functions with well-defined structural properties, proves to be an effective and reliable tool for algorithm analysis in cases where full enumeration is infeasible. The obtained results represent a significant step toward a deeper understanding of algorithmic behavior in large solution spaces, supporting the generalization of earlier observations and providing an empirical foundation for further research in adaptive heuristics. As a result, the presented analyses not only confirm the scalability of the phenomena observed in the small function space but also offer a methodological research template that can be extended to other function classes and even larger dimensions in future studies.

\subsection{Conclusions and future research directions}  

The conducted research provides an innovative perspective on the impact of algebraic operations on test functions in permutation-based optimization algorithms. The results indicate that both the addition and subtraction of binary functions within the c.u.p. structure lead to systematic correlations in algorithmic performance, challenging the classical assumptions of the NFL theorem in a constrained, local setting. Previous approaches to heuristic optimization analysis have not accounted for structural dependencies within the test function space, relying instead on the assumption of complete randomness in outcomes, which suggests the necessity of revisiting these concepts.

The issue of function structure influence on algorithm performance has been previously discussed in the works of Clerc \cite{3} and Whitley \cite{11uuu}, yet their analyses focused primarily on the properties of individual test functions and their impact on the exploration process within the solution space. Whitley pointed out fundamental limitations of heuristics related to the landscape of the objective function but did not consider algebraic transformations as an analytical tool for algorithm assessment. Clerc, in his research, emphasized structural dependencies in optimization problems but did not relate them to systematic modifications of test functions. It is only through our research that we demonstrate how algebraic operations can introduce a new dimension of analysis, enabling the identification of performance patterns in algorithms that have previously eluded traditional evaluation methods.

Similar attempts to challenge the generality of the NFL theorem can be found in the work of Kimbrough et al. \cite{10ak}, who showed that specific problem classes, such as constraint optimization problems handled by the Feasible-Infeasible Two-Population (FI-2Pop) Genetic Algorithm, can violate the NFL assumptions. Their research focused on the impact of constraints on algorithm efficiency, demonstrating that problem structure can favor certain heuristic methods. Our analyses extend this line of research, showing that structural dependencies in test functions also lead to non-uniform distributions of algorithm efficiency within structured benchmarks. The results indicate that certain algorithms exhibit systematic performance patterns on algebraically related functions, suggesting that the assumption of complete performance randomness may be flawed for real-world optimization problems. 
This aligns with findings in the paper \cite{10aa}, where a MAX-2-SAT problem with an objective value-permutation that does not conform to a standard MAX-2-SAT problem is presented. This result shows that MAX-2-SAT is not c.u.p.  for the case n=3, with proof hints also provided for n=4. 
Previously, the MAX-SAT problem was explored in works such as \cite{9}, \cite{11}, and \cite{11aa}, further underscoring how problem-specific structures can lead to deviations from general theoretical assumptions.

Future research should focus on exploring the impact of multiplicative algebraic operations on test functions, which may reveal additional layers of structural dependencies determining the performance of optimization algorithms. Extending analyses to higher-order transformations, including polynomial or nonlinear 
combinations of test functions, could further refine algorithm classification concerning their resistance to specific problem types. Additionally, the application of machine learning methods to predict heuristic performance based on the algebraic properties of test functions could lead to the development of new adaptive optimization strategies.

Ultimately, this study opens new perspectives in heuristic optimization, providing an algorithm analysis methodology that transcends standard statistical randomness models and captures deep structural dependencies within test function spaces. The ability to more precisely match algorithms to specific problem classes implies that future research could more accurately define the limits of the NFL theorem's validity and identify scenarios in which structural dependencies within test functions determine the selection of an optimal heuristic.

\section{Implications}

The results reported in this paper indicate that permutation structure and objective reformulation can systematically influence algorithm behavior. This is particularly relevant for inference and optimization procedures that rely on relabeling, resampling, or permutation-based search. In such settings, evaluation order 
and symmetry can shape early-stage information acquisition.
Below, we focus on implications for statistics and evolutionary computing. The discussion emphasizes structurally restricted function classes (including c.u.p.) and the arithmetic structure of objectives, which can induce consistent re-rankings among permutational strategies.

\subsection{Implications for Statistics}

The No Free Lunch theorem carries significant consequences for statistics, particularly in the design and application of algorithms for specific analytical tasks. This study focuses on permutational algorithms - a class of methods that operate by permuting elements within a dataset or objective function-and investigates their behavior under conditions that violate the assumptions of the NFL theorem. In particular, it analyzes function subsets c.u.p. and their impact on algorithmic performance. Below, the findings are discussed in the context of concrete statistical applications:

\subsubsection{Permutational Significance Tests in Multivariate Statistics}

In multivariate statistics, especially in fields such as genomics or neuroimaging  \cite{4a},\cite{16a},\cite{10b} hypothesis testing is often used to examine group differences. Traditional parametric tests such as Student's t-test \cite{1818} or ANOVA  \cite{1010}, rely on assumptions of normality and homogeneity of variance, which are frequently violated in practice. The results presented in this study suggest that for data satisfying the c.u.p. condition (i.e., when the distribution of results is closed under permutation), permutational algorithms may offer a more robust alternative. In a permutation test, the null hypothesis is assessed by repeatedly shuffling group labels and recalculating the test statistic to generate an empirical distribution.

Our experiments with 24 permutational algorithms $a1$ - $a24$ on artificial function sets revealed substantial performance differences across algorithms within constrained problem subsets. For example, for test function $f14$ (constructed as the sum of two simpler functions, $f5$ and $f10$, where $f5$ represents a smooth unimodal convex structure and $f10$ introduces discontinuities and noise), one algorithm $a4$ located the optimum significantly faster than when each component function was evaluated independently. This indicates that leveraging the structural properties of the c.u.p. set - such as additive effects that create a richer and more discriminative optimization landscape - enhanced sensitivity to real differences. Similarly, in empirical permutation tests, an algorithm optimized for a specific cost function (e.g., sum of squared differences) can more effectively detect subtle discrepancies between groups when data exhibit an ordered structure. Employing such tailored permutational approaches increases the power of the test (the ability to detect true effects) while simultaneously reducing the risk of Type I errors.

\subsubsection{Optimization of Regression Estimators via Permutations}

Regression analysis  \cite{9a},\cite{10x}  often encounters difficulties caused by outliers or heteroscedasticity, which undermine the assumptions of classical Ordinary Least Squares (OLS) methods. While robust methods like Least Absolute Deviations (LAD) exist, they may also be inefficient under certain data configurations. 
This study proposes a framework for optimizing regression estimators through a focus on permutational data subsets and an analysis of the arithmetic properties of objective functions. Our results indicate that violations of the NFL theorem on c.u.p. sets can be leveraged to improve parameter estimation.

To illustrate, consider financial data where income distributions exhibit ''heavy tails'' (substantial deviation from normality). Rather than using classical OLS, one can apply a permutational algorithm that rearranges residuals to minimize a suitably constructed loss function that accounts for this structure (e.g., a cost function that penalizes large errors more heavily). Our experiments confirm the validity of this approach: performance on a composite function (e.g., $f14 = f5 + f10$) was not additive with respect to the results on its components. In other words, the cumulative effects of two features can create an optimization landscape that favors certain methods more than would be expected from their individual behaviors. This synergy effect translates to more accurate and stable regression estimates in cases where classical assumptions fail. A permutational strategy thus yields more precise and interpretable models, particularly for datasets with mixed quality or numerous outliers.

\subsubsection{Permutational Bootstrap Methods}

The bootstrap technique  \cite{2bb},\cite{13c},\cite{10b} is a fundamental tool for estimating sampling distributions and constructing confidence intervals. However, traditional bootstrap procedures can be computationally expensive and may not fully exploit data structure. Our findings show that permutation-based bootstrap methods can be optimized for c.u.p. subsets, reducing computational complexity while maintaining statistical accuracy. That is, violations of NFL assumptions in restricted problem sets allow the algorithm to focus on the most informative permutations.

For instance, in time-series analysis, block bootstrap methods sample contiguous blocks of data to preserve temporal dependencies. These can be enhanced using permutational algorithms that respect the sequential structure of the data (i.e., avoiding the disruption of natural groupings). Our analysis shows that certain permutational algorithms (e.g., the $a2$ - $a4$ - $a6$ group characterized by systematic search) can generate representative permutations with fewer iterations, thus reducing computational load in the bootstrap context. This enables reliable estimation of confidence intervals for complex datasets (e.g., high-frequency financial 
series or environmental data) with significantly lower computational cost.

\subsubsection{Permutational Tests in Mixed-Effects Models}

Mixed-effects models \cite{9xx}, used to analyze hierarchical data structures (e.g., students in classrooms, patients in hospitals), pose challenges when testing the significance of fixed and random effects due to their complexity and sensitivity to distributional assumptions. Our findings suggest that permutation tests can be effectively applied in mixed-effects contexts by using c.u.p. sets to construct empirical distributions of test statistics.

The key idea is to permute data  within natural groups (e.g., within each classroom or hospital) rather than across the entire dataset. This preserves the inherent dependency structure among observations and avoids violating model assumptions. For example, in an educational study assessing a new teaching method, the significance of classroom-level effects can be tested by permuting student assignments within classes. Our experiments revealed similar benefits from respecting structure: functions with similar properties formed homogeneous clusters (e.g., $f1$, $f2$, and $f5$ exhibited similar performance profiles), supporting the consistent 
behavior of certain algorithm groups. This suggests that restricted group-wise permutations yield more reliable test results in mixed models than traditional methods, eliminating the need for strict parametric assumptions.

\subsubsection{Analysis of Nonlinear Dependencies in Multivariate Data}

Identifying nonlinear dependencies  \cite{3a},\cite{8b},\cite{10xyy},\cite{22xy} among variables is a common challenge in multivariate analysis. Standard association measures such as Pearson's correlation may fail to detect such relationships when the dependence is non-linear. Permutational algorithms introduced 
in this study offer a powerful tool for exploring nonlinear dependencies by deliberately introducing permutations and observing the effects on a target function (e.g., mutual information or distance metrics). This approach can uncover hidden patterns in the data.

In our experiments, applying permutations to combine or modify test functions revealed dependencies not evident in the individual analyses. For instance, summing two functions $f5$ and $f10$ to form $f14$ changed the optimization landscape such that algorithms responded differently than when evaluating each component separately. This shift indicates nonlinear interactions between the components-analogous to statistical data where a significant change in a criterion function after permuting a variable implies a nonlinear relationship with other variables. Furthermore, our correlation and PCA analyses demonstrated  (see Section 8.1) that introducing such nonlinear transformations (e.g., summing or subtracting functions) generated new directions of variability in algorithm outputs. 
Similarly, permutation-based procedures can detect nonlinear inter-variable relationships that linear association measures may miss.
As shown in \cite{678}, nonlinear short-term forecasting on empirical time series can reveal signatures of deterministic chaos and separate them from sampling/measurement error and externally induced noise.
For environmental predictors (e.g., temperature and precipitation), we probe nonlinear dependence by shuffling variables and monitoring the change in an objective criterion (e.g., mutual information).

\subsubsection{Bayesian Statistics}

Bayesian statistics is based on the iterative updating of probability distributions as new data become available \cite{2xy},\cite{4abc},\cite{4zw},\cite{11xyz}. Unlike frequentist approaches, Bayesian methods incorporate prior knowledge in the form of a prior distribution, which is then updated via the likelihood function 
to obtain the posterior distribution. This approach allows for considerable modeling flexibility - from analyses based on small samples to complex hierarchical models - but the effectiveness of Bayesian methods strongly depends on the appropriate selection of prior distributions and the efficiency of algorithms for exploring the parameter space (e.g., the efficiency of Markov chain Monte Carlo (MCMC) sampling \cite{GOSK},\cite{4xx}).   %(e.g., performance of Monte Carlo Markov Chains, MCMC  \cite{4xx}).
The results obtained in this study suggest that conscious incorporation of problem structure (e.g., restricting the function space to specific classes) may favor certain analytical strategies over others. In other words, using additional (structural) knowledge - analogous to the role of the prior distribution in the Bayesian 
paradigm - can break the universal assumptions of the No Free Lunch theorem and improve inference outcomes for specific problem classes.

The permutational strategies proposed in this study can enhance Bayesian modeling by more effectively matching model assumptions to the actual structure of the data. 
When data are skewed or heavy-tailed, Gaussian assumptions can be brittle, so robustness is improved by using heavy-tailed priors and/or likelihoods (e.g., Student-t).
For example, when data exhibit clear skewness or heavy tails (e.g., due to outliers or missing values), standard prior assumptions (such as the classical 
Gaussian prior \cite{VV}) may be inadequate. Instead, one should employ priors that reflect the observed structure - such as asymmetric or heavy-tailed distributions - to improve model fit and interpretation. Our analyses confirm this need: the performance distributions of algorithms after function modifications show extreme variation - values range nearly across the full spectrum from close to 0 to 1, with some algorithms achieving very low efficiency while others perform very well. Such spread (presence of outliers, visible in Figure 10) indicates data heterogeneity and corresponds to heavy-tailed distributions of performance. Therefore, in Bayesian analysis, one should use prior distributions capable of capturing this variability (e.g., heavy-tailed rather than normal priors) to avoid misleading assumptions and improve inference reliability.

\vspace{0.4cm}
\textit{Permutational MCMC Strategies and Parameter Grouping}\\

Our correlogram and hierarchical-clustering results reveal clear structural clusters in the benchmark (see Figures~2 - 3; see also Figures~4 - 7 for transformed benchmarks).
Such structure suggests strong dependencies; in Bayesian inference, blocked (grouped) MCMC updates and the choice of scan order can improve mixing and reduce autocorrelation \cite{GOSK},\cite{4xx}.
For instance, functions $f1$, $f2$, and $f5$ form a homogeneous group (Figure~3).
The composite function $f14=f5+f10$ shows that algebraic structure can change the performance landscape: empirically, $M(a4,f14)\neq M(a4,f5)+M(a4,f10)$.
Finally, some algorithms behave more heterogeneously (e.g., the $a15$ - $a17$ group), while others show weak similarity to most methods (e.g., $a9$), supporting the view that the benchmark decomposes into strongly related groups.

\vspace{0.4cm}
\textit{Applications in Hierarchical Models}\\

Our correlograms and dendrograms (see Figures~2 - 7) show that performance dependence is not uniform across the benchmark.
Instead, it concentrates within coherent groups.
A similar situation arises in Bayesian inference under strong posterior dependence.
In such cases, blocked (grouped) updates and the choice of scan order can improve mixing and reduce autocorrelation \cite{GOSK}.
Motivated by this analogy, we propose a preprocessing step.
We use correlation-based clustering to define parameter blocks before running MCMC.
See the $f14=f5+f10$ example above for a concrete illustration of how structural coupling changes efficiency.

\vspace{0.4cm}
\textit{Monte Carlo Simulations and Algorithmic Efficiency}\\

Monte Carlo experiments (see Section~8.4) show that sensitivity to sampling order persists for larger problem sizes $n=10, 30, 50, 100$ and depends strongly on function class.
For $n=30$, the gap between the best and worst sampling orders is 22 - 30 percentage points; mean efficiencies are $63.4\%$  for balanced functions, $49.5\%$  for strongly unbalanced functions, and $70.6\%$  for symmetric functions.
This empirical pattern motivates importing structure-aware sampling ideas into Bayesian computation, where blocked updates and scan order are known to affect mixing and autocorrelation  \cite{GOSK}.

\subsection{Implications for Evolutionary Algorithms}

This study highlights that NFL equalities rely on specific conditions. In particular, they assume uniform averaging over the function class and permutation-closure. When these conditions are not met, search strategies can differ systematically in a benchmark-dependent way \cite{4bbb},\cite{8},\cite{9},\cite{10aa},\cite{100bb},\cite{11},\cite{11aa},\cite{12}. This directly affects how evolutionary algorithms should be designed and applied. C.u.p.\ function classes indicate when NFL-style average-case equivalences hold. However, our structured benchmarks and algebraic objective recombinations produce consistent re-rankings across algorithms. 
They also reveal non-additive efficiency on composite objectives, especially when objective structure is exploited (e.g., symmetry and compositional structure). PCA analyses (see Section~8.1) confirm clear shifts in data structure after function transformations. Even simple additions or subtractions substantially reorganize function-algorithm relationships. The key practical guidelines for evolutionary algorithms are discussed below.

\subsubsection{Designing Operators}

Both the correlations of algorithm results and the hierarchical clustering analysis indicate that the design of genetic operators (mutation, crossover, selection) should take into account the structure of the problem \cite{4bbb},\cite{100bb} (see, e.g., an applied example in \cite{7az}).
In the conducted experiments, groups of algorithms with nearly identical behavior were observed - for example, algorithms $a1$ and $a2$ exhibited an almost perfect correlation of results across the entire set of test functions. Such pairs (differing only by minor heuristic changes, e.g., the order of search) achieved similar outcomes, which suggests that certain modifications of operators do not introduce significant changes in performance. In other words, if an evolutionary operator does not exploit the specific structure of the objective function landscape, the algorithm may turn out to be redundant in relation to another with a similar operational scheme.

On the other hand, the results highlight the need to design operators that are robust to algebraic changes in the objective. In Data2, negative deviations dominate and often form block-structured regions (e.g., a pronounced drop for $f2$ spanning roughly $a9$ - $a24,$ and mostly negative values for $f9$ over $a1$ - $a16,$ followed by a sign reversal around $a19$ - $a22$), with severe local degradations such as $f5$ - $a21$ and $f5$ - $a22.$ Positive deviations are present but typically localized, including $f8$ - $a1$ and gains on $f12$ for $a2$ (also $a11$ and $a22$), as well as multiple gains on $f14$ (e.g., $a4,$ $a6,$ $a14,$ $a20$). Panel 8b) (Figure 8) is more heterogeneous and contains larger contiguous positive regions, indicating stronger pair-dependent re-ranking; it should be interpreted as the net effect of algebraic recombination rather than a purely subtractive perturbation. These patterns motivate operator designs that maintain search effectiveness under structured objective recombinations.

Consequently, in the design of operators for evolutionary algorithms \cite{4bbb},\cite{10ccg},\cite{10cch}, it is worthwhile to consider the algebraic properties of the objective function. Our results indicate that the structure of the problem (e.g., permutation symmetries or objective additivity) can affect the effectiveness of operators - for example, adding two functions may reduce the symmetry of the landscape and facilitate the algorithm’s ability to distinguish good from bad solutions. This suggests that crossover operators that preserve favorable permutation segments or mutation operators aligned with the representation may be beneficial for convergence. For instance, in permutation problems such as the traveling salesman problem \cite{2222}, a customized crossover that retains promising sub-permutations of cities could preserve good partial structures while still exploring new configurations. Similar conclusions follow from the analysis of covariance of algorithm results: strongly cohesive groups emerge (e.g., $a1$ - $a4$ and $a5$ - $a7$), while some algorithms exhibit more context-dependent behavior under transformations (e.g., $a15$ - $a17$), and others show irregular transitions between positive and negative covariance patterns (e.g., $a20$ - $a22$). These observations suggest that operators promoting stable performance across algebraic transformations of the objective should be prioritized, especially in variable or multi-component optimization settings.

\subsubsection{Multi-Objective Evolutionary Algorithms (MOEA)}

Although the present study evaluates permutation-based benchmarks in a single-objective setting using an efficiency-oriented performance measure, the findings are informative for multi-objective evolutionary algorithms (MOEAs) (see e.g., \cite{2asd},\cite{4yyy}), where objective criteria are frequently aggregated, decomposed, or otherwise algebraically recombined during problem formulation and analysis. In MOEA studies, approximation sets are commonly assessed with indicators such as hypervolume and IGD\cite{Yi}.

Our experiments show that simple algebraic recombinations of objective functions (addition/subtraction) can significantly affect algorithm performance even within the same c.u.p.\ function space. This effect is confirmed by a one-way ANOVA over the element-wise performance values across the three datasets (Data1, Data2, Data3), which yields $F=110.68$ with $p = 3.59965 \times 10^{-44}$ (Table~5). Tukey’s post-hoc test (Table~6) shows statistically significant differences between Data1 and Data2 (mean difference $=0.32639$, $p=0.0000$) and between Data1 and Data3 (mean difference $=0.33532$, $p=0.0000$), while Data2 vs.\ Data3 is not significant (mean difference $=0.00893$, $p=0.93556$). Thus, the original benchmark set differs markedly from both algebraically recombined sets, whereas the two recombined datasets are statistically indistinguishable under this test.
The transformed datasets were generated from the original c.u.p.\ functions using algebraic operations, while preserving permutation-closure. Nonetheless, function-algorithm relationships changed substantially (see Figure~8). Data2 was generated via a mixture of sums and differences (e.g., $f6 = f2 + f5$ and $f5 = f14 - f10$). In contrast, Data3 was constructed primarily from differences, but it still includes several sums (e.g., $f2 = f6 - f5$ and $f12 = f11 + f2$). These observations suggest a practical implication for MOEAs. The specific algebraic form used to combine criteria may change the effective difficulty of the search. It may also reorder which strategies appear most effective for a given formulation. For example, this includes aggregating components vs.\ keeping them separate, or reformulating objectives via structured recombinations. Depending on the formulation, such changes may also alter dominance relations in the objective space.
Finally, the algorithm - algorithm correlation matrices (see Figures 2/6) and the covariance-based hierarchical clustering heatmaps (see Figures 3/7) indicate that algorithm families differ in their sensitivity to such transformations. Strongly cohesive groups (e.g., $a1$ - $a4$ and $a5$ - $a7$) exhibit highly aligned behavior. Other groups show more context-dependent patterns under subtraction-based modifications (e.g., $a15$ - $a17$). Some subgroups display inverse covariance relative to the dominant clusters (e.g., $a10$ - $a12$). For MOEA design, this motivates using operators and selection mechanisms that remain effective under structured changes of the objective formulation. When permutation symmetries are present, it also motivates incorporating symmetry-aware variation or evaluation. This can reduce redundant search over equivalent representations.

\subsubsection{Hyperparameter Optimization with Permutational Algorithms}

Hyperparameter optimization (HPO) is a central task in machine learning. The goal is to find configurations that maximize model performance under limited evaluation budgets \cite{1aa},\cite{2x2},\cite{4jjj}. In practice, evolutionary and population-based search strategies are frequently considered for large or discrete design spaces. This includes settings related to neural architecture design \cite{10yui},\cite{12uuu}. It also includes structured configuration choices in discrete or permutation-based spaces \cite{5},\cite{100bb},\cite{11z2}.
A particularly relevant HPO regime occurs when part of the configuration is naturally represented as a \emph{permutation}. In this case, the order of components matters and symmetry is present.   

In the experimental setup reported above, we sought a controlled empirical basis for understanding how a purely permutational search structure affects optimization efficiency. We analyzed 24 permutation-based algorithms that share an identical mechanism. They differ only in the order in which they evaluate a fixed set of candidate points (Table~2). This design isolates the effect of \emph{evaluation order} as the only algorithmic degree of freedom. The performance measure used in the experiments is the normalized efficiency index $E_{a_i,f_j}$ (see Section 4.2). It is defined by the trial at which the known optimum is first reached - Equation~(1). This index directly captures how early useful information is acquired during the search.

Two observations are particularly informative for HPO with permutation structure. First, correlation and clustering analyses reveal strongly cohesive families of algorithms whose performance profiles are highly aligned. This includes near-perfect similarity for $a1$ - $a2$, and a strong relationship for $a12$ - $a16$. This indicates that small changes in permutational evaluation order can lead either to practically redundant behavior (high correlation) or to distinct exploration/exploitation trajectories (weak or negative correlations). In HPO practice, this supports using correlation- and cluster-based diagnostics to avoid maintaining multiple algorithm variants that are effectively interchangeable. It also supports selecting genuinely complementary search orders when designing ensembles or hybrids.

Second, the study shows that algebraic recombinations of objective functions can systematically reshape function - algorithm relationships. This happens even within the same c.u.p.\ function space. The benchmark transformations Data2 and Data3 are constructed from Data1 by structured sums and differences. For example, in Data2: $f13=f5+f9$. For example, in Data3: $f2=f6-f5$. These transformations preserve permutation-closure. Differential heatmaps (Figure~8) show broad, structured performance shifts. They include block-like degradations and localized gains. ANOVA with Tukey post-hoc tests also confirms a highly significant dataset effect (see Table~5 - 6). For HPO, this provides a concrete caution. Seemingly simple objective reformulations (e.g., adding penalty terms, changing how criteria are combined, or replacing a base objective with a structured recombination) may change the effective search landscape. They may also reorder which strategies are most efficient, even when the underlying symmetry class remains unchanged.

These results motivate permutation-aware HPO designs that explicitly exploit symmetry. This reduces redundant exploration of equivalent representations. They also motivate prioritizing robustness to structured objective reformulations, because objective recombinations can substantially change 
which search orders are efficient (see Figure 8; see also Tables 5 - 6). Finally, empirical dependency analyses (correlation and clustering) can be used to select non-redundant algorithmic variants and assemble more complementary search components.

\subsubsection{Quadratic Assignment Problem (QAP)}

The Quadratic Assignment Problem (QAP) \cite{RSL} is a classical NP-hard permutation problem. The goal is to assign $n$ facilities to $n$ locations while minimizing an interaction cost. The objective is typically expressed as a sum of products of pairwise flows and the corresponding distances induced by the assignment permutation. The search space has size $n!$, which makes QAP challenging for general-purpose heuristics.
From the NFL viewpoint, no optimizer is uniformly best when performance is averaged over all objective functions  \cite{8},\cite{9},\cite{10},\cite{10x},\cite{10aa},\cite{16ro}. Therefore, systematic advantages must be tied to structural regularities of the targeted problem class. Our controlled c.u.p.\ benchmark supports this practical interpretation. Correlation and clustering reveal families of near-redundant search orders (see Figures~2 - 3). They also reveal outlier behaviors, such as $a19$, $a21$, and $a23$, which deviate from the dominant clusters and are consistent with specialization to particular landscape types rather than broad robustness.
Algebraic reformulations of the objective substantially reshape function-algorithm relationships (see Figure~8). The resulting delta patterns show that the same structured recombination can penalize broad algorithm groups while benefiting narrower subsets. This is consistent with local performance inequalities emerging under structured transformations, even though the global NFL statement remains intact for the full function class. In addition, structural additivity of objective components does not imply additive search effort. In our experiments, even when a function can be written as a pointwise sum (e.g., $f14$ arising from $f5+f10$ in the binary c.u.p.\ construction), the performance measure $M$ need not satisfy $M(a,f_i+f_j)=M(a,f_i)+M(a,f_j)$. Such non-additivity is compatible with the interpretation that composition can change symmetry and discriminability in the landscape.
These observations motivate QAP-oriented algorithm design principles. QAP costs decompose into many pairwise interactions, so operators and models can explicitly exploit permutation structure and interaction patterns. At the same time, they should remain robust to structured objective reformulations, because even simple algebraic recombinations can re-rank search orders. One principled direction is to use assignment-aware probabilistic models, such as doubly stochastic matrix (DSM) models within EDAs for QAP \cite{10sfj}.

\subsection{Conclusion}

No Free Lunch theorems are average - case statements. Under uniform averaging over a finite
function class and non - revisiting search, all optimizers have identical expected performance \cite{12},\cite{28x}.
In the sharpened finite - domain formulation, under uniform weighting this equivalence holds if and only if
the function class is closed under permutation (c.u.p.) \cite{9}. Our fully enumerated $n=4$ baseline is
c.u.p.\ by construction. Consistently, all 24 evaluation orders listed in Table~\ref{EVO:tab-24-compact} attain the same mean 
efficiency in Data1 (Table~\ref{tab23}). NFL constrains the uniform average over the class; it does not imply equality on each
individual objective.
Our contribution is to characterize what remains informative beyond this average. Structure - preserving
algebraic objective reformulations (sums and differences) systematically reorganize function - algorithm
dependencies. They reveal stable families of near - equivalent evaluation orders and transformation - sensitive
variants. We also observe non - additivity of search effort: performance on $f_i \pm f_j$ cannot be inferred
from performance on $f_i$ and $f_j$ alone. Practically, objective formulation and benchmark construction
define the effective testing regime and should be treated as part of algorithm design rather than neutral
preprocessing. Beyond optimization benchmarks, the same NFL lens applies whenever the task distribution is shaped by
symmetry or structural restrictions. This includes permutation-based inference and stochastic simulation.
In such regimes, the induced problem family can make evaluation order and objective coupling practically
consequential.

The article has the potential to influence further research in the field of the mathematical foundations of heuristic optimization and is recommended for scientists, engineers, and practitioners involved in optimization and evolutionary algorithms.\\

The data supporting the findings of this study are available from the corresponding author upon reasonable request.\\

All numerical computations, data processing, and statistical visualizations presented in this study were performed using \textbf{MATLAB R2023b} on a DELL computer (system specification: Windows 10 Home, version 22H2).

\section*{Acknowledgements}

The author gratefully acknowledges Professor S\l awomir T. Wierzcho\'n (Institute of Computer Science of Polish Academy of Sciences, Warsaw, Poland)  for the initial inspiration and encouragement that helped shape this research direction, for drawing the author’s attention to the work of M. Clerc, and for thoughtful comments and suggestions on both the substantive content and the editorial presentation of the manuscript, which improved its clarity.

\renewcommand{\refname}{
\end{document}